\documentclass[10pt,twocolumn,letterpaper]{article}

\usepackage{cvpr}              % To produce the CAMERA-READY version
\usepackage{times}  % DO NOT CHANGE THIS
\usepackage{helvet}  % DO NOT CHANGE THIS
\usepackage{courier}  % DO NOT CHANGE THIS
\usepackage[hyphens]{url}  % DO NOT CHANGE THIS
\usepackage{graphicx} % DO NOT CHANGE THIS
\usepackage{natbib}  % DO NOT CHANGE THIS AND DO NOT ADD ANY OPTIONS TO IT
\usepackage{caption} % DO NOT CHANGE THIS AND DO NOT ADD ANY OPTIONS TO IT
\usepackage{algorithm}
\usepackage{algorithmic}

\usepackage{subcaption} % 导言区添加
\usepackage{pifont}
\usepackage{amsmath}
\usepackage{amsfonts}
\usepackage{multirow}
\usepackage{mathabx}
\usepackage{xcolor}
\usepackage{caption}
\usepackage[table]{xcolor}
\usepackage{subcaption}
\usepackage{newfloat}
\usepackage{listings}
\usepackage{soul}        % 导言区
\sethlcolor{yellow}
\usepackage[accsupp]{axessibility}  %
\soulregister\cite7
\soulregister\ref7
\soulregister\citet7
\soulregister\citep7
\definecolor{cvprblue}{rgb}{0.21,0.49,0.74}
\usepackage[pagebackref,breaklinks,colorlinks,allcolors=cvprblue]{hyperref}

\def\paperID{40499} % Enter the Paper ID here
\def\confName{CVPR}
\def\confYear{2026}

\title{Weakly-Supervised Referring Video Object Segmentation\\ through Text Supervision}

\author{
Miaojing Shi$^{1}$, \quad Jun Huang$^{1}$, \quad Zijie Yue$^{1}$\thanks{Corresponding author.}, \quad Hanli Wang$^{1}$ \\ 
$^{1}$College of Electronic and Information Engineering, Tongji University\\
{\tt\small \{mshi, jhuang, zijie, hanliwang\}@tongji.edu.cn}
}

\begin{document}
\maketitle

\begin{figure*}
\centering
\captionsetup{aboveskip=10pt, belowskip=5pt}
\includegraphics[width=\linewidth]
{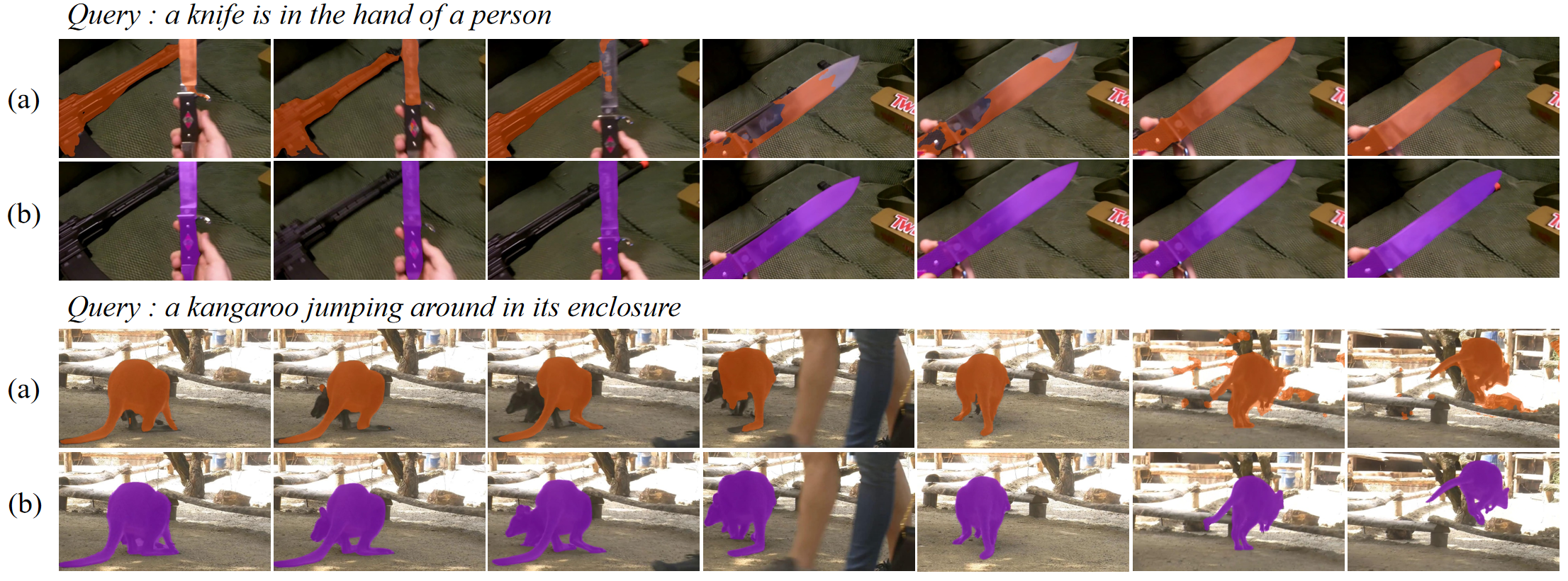}
   \caption{Visualization results on Ref-YouTube-VOS. (a) OCPG (point-supervised method)~\cite{ocpg}, (b) our proposed WSRVOS. }
   % WSRVOS effectively ensures the global temporal consistency of segmentation results, thereby reducing objects misalignment.Qualitative results of the proposed WSRVOS across video frames. GT: ground truth.
\label{qualitative_results}
\vspace{-5mm}
\end{figure*}

\begin{abstract}
Referring video object segmentation (RVOS) aims to segment the target instance in a video, referred by a text expression. Conventional approaches are mostly supervised learning, requiring expensive pixel-level mask annotations. To tackle it, weakly-supervised RVOS has recently been proposed to replace mask annotations with bounding boxes or points, which are however still costly and labor-intensive. In this paper, we design a novel weakly-supervised RVOS method, namely WSRVOS, to train the model with only text expressions. Given an input video and the referring expression, we first design a contrastive referring expression augmentation scheme that leverages the captioning capabilities of a multimodal large language model to generate both positive and negative expressions. We extract visual and linguistic features from the input video and generated expressions, then perform bi-directional vision-language feature selection and interaction to enable fine-grained multimodal alignment. Next, we propose an instance-aware expression classification scheme to optimize the model in distinguishing positive from negative expressions. Also, we introduce a positive-prediction fusion strategy to generate high-quality pseudo-masks, which serve as additional supervision to the model. Last, we design a temporal segment ranking constraint such that the overlaps between mask predictions of temporally neighboring frames are required to conform to specific orders. Extensive experiments on four publicly available RVOS datasets, including A2D Sentences, J-HMDB Sentences, Ref-YouTube-VOS, and Ref-DAVIS17, demonstrate the superiority of our method. Code is available at \href{https://github.com/viscom-tongji/WSRVOS}{https://github.com/viscom-tongji/WSRVOS}.

\end{abstract}    
\section{Introduction}
Referring video object segmentation (RVOS) aims to generate the mask prediction for the target instance in a video, referred by a text expression. It can benefit many practical applications such as text-based video editing, surveillance and human-computer interaction. Gavrilyuk~\textit{et al.}~\cite{firstRVOS} introduced this task and proposed an encoder-decoder structure to generate the segmentation mask by convolving visual features with dynamic filters obtained from linguistic features. Subsequently, many methods were introduced to fuse multimodal features and predict the segmentation results, of which single-stage methods~\cite{RefVOS, CMSA} directly fuse visual and linguistic features to predict masks; two-stage methods~\cite{AAMN, clawcranenet} first generate mask candidates and then select the one that best matches the input expression.
%in either single-stage~\cite{Urvos, RefVOS, CSTM, CMSA} or two-stage manner~\cite{AAMN, rethinking, clawcranenet}. 
Recently, due to the effectiveness of transformer architecture in computer vision~\cite{ViT,swint}, query-based RVOS approaches~\cite{ReferFormer,MTTR, pansemantic} have become the mainstream. 

Despite the significant progress of existing RVOS methods, they heavily rely on expensive pixel-level mask annotations for supervision, hence limiting their applications. A promising direction is to solve RVOS with weakly-supervised learning, \ie, leveraging bounding box or point annotations on target instances for training supervision. However, such annotations remain costly due to repetitive manual labeling across frames.
% However, such annotations still require manual efforts for many video frames and thus remain costly and labor-intensive. 
To alleviate it, we aim to train the model to locate target instances based on text expressions solely.
% \hl{To alleviate this burden, recent works have explored weakly-supervised RVOS as an alternative. Most of them enhance text supervision with coarse spatial guidance like bounding boxes or point annotations. However, these additional supervisions remain comparatively costly due to the manual need for repeated annotations across frames.
% % While weakly-supervised RVOS methods has recently emerged as an alternative, most such methods depend on additional supervision such as bounding boxes or point annotations, still requiring manual effort.
% To address this limitation, we introduce the weakly-supervised RVOS using only text supervision, a more practical setting in which any kind of manual annotation is unavailable during training.
% % Instead, the model must learn to localize target instances solely based on text expressions.several works have explored weakly-supervised RVOS in which the mask annotations are replaced with bounding boxes or point annotations for training supervision. However, such annotations still require manual efforts for every video frame and thus remain costly and labor-intensive. To overcome this limitation, we aim to train the model to learn to localize target instances solely based on text expressions.
This presents several challenges: the first is the heterogeneity between visual and linguistic features, making it difficult for the model to align high-level semantic information between them; second, occlusions, motion blur, and the temporal dynamics of videos further complicate this alignment.

Recently, the advent of multimodal large language models~\cite{Blip2,llava,Qwen-VL,qwen3} (MLLMs) paves a promising avenue to address above challenges. In particular, the captioning capabilities of MLLMs, such as Qwen3-VL~\cite{qwen3}, indeed provide a new way to overcome the supervision insufficiency in the weakly-supervised RVOS task.
By leveraging MLLMs to generate diverse text expressions to describe various aspects and contexts of visual content, we can obtain rich and diverse supervision signals to facilitate effective and fine-grained alignment between visual and linguistic features, thereby improving the model’s ability to locate the target instance under weak supervision.

% In terms of the challenge of existing weakly-supervised referring segmentation methods' weak ability to align multimodal features, we consider to solve this problem from one perspective: the diversity of text supervision. 
% By providing more text expressions as supervision for each target instance, the model can learn to align the visual attributes of the referred object with diverse referring expressions. 

% which are incorrect yet semantically plausible with the expression of the video frame.

%given the original videos and text expressions. 
%These expressions are regarded as extra positive expressions since each of them refer to a target instance in the input videos like the original texts. 

% Inspired by TRIS~\cite{TRIS}, we find regions within the video frames that exhibit high similarity to the generated positive and negative referring expressions, and apply an image-level classification loss to guide the model learn how to locate the referred targets. 
% Since video frames often contain extensive and redundant visual features, directly aligning them with text expressions can lead to inaccurate cross-modal associations. 
% unlike previous works that directly perform cross-modal fusion for target localization,In this paper, we introduce the first end-to-end weakly-supervised RVOS framework (WSRVOS).
In this paper, we introduce an end-to-end weakly-supervised RVOS framework (WSRVOS) that relies solely on text supervision.
Our first contribution lies in a \emph{Contrastive Referring Expression Augmentation scheme}: given a video and its original referring expression, we leverage a MLLM to generate additional positive expressions by explicitly enriching the original expression with details related to visual appearance, actions, and inter-instance relations. In parallel, we also generate more expressions that are semantically plausible yet inconsistent with the original expression, serving as negatives to facilitate more discriminative multimodal alignment.
Next, we focus on leveraging the text supervision to train a robust weakly-supervised RVOS model. We first extract visual and linguistic features from the given video and the generated expressions.
Since videos often present temporally dynamic and semantically diverse content, their visual features inevitably include redundant or irrelevant information that fails to correspond to the referring expression. The expressions themselves may also include auxiliary words (\eg prepositions) that are unrelated to the video content. Therefore, we apply a \emph{Bi-directional Vision-Language Feature Selection module} to select visual and linguistic features that are highly relevant to each other. 
These selected features are then leveraged to enable fine-grained multimodal alignment, termed as an \emph{Instance-aware Expression Classification scheme}. It performs proposal aggregation and expression matching, inspired by multiple instance learning \cite{MIL}, to let the model distinguish positive expressions from negative expressions in videos.
Moreover, we propose a \emph{Positive-Prediction Fusion strategy} to enable supervision by integrating predictions from positive expressions. By fusing these predictions, we produce reliable pseudo-masks that serve as effective supervision signals to facilitate more accurate localization of the target instance. 
Finally, we design a \emph{Temporal Segment Ranking constraint} to optimize the overlap values between mask predictions of temporally neighboring frames, encouraging higher overlaps for closer frames than for distant ones.
% employs a ranking-based loss over adjacent and longer-range frame pairs, thereby strengthening the temporal coherence of the segmentation results.to optimize the consistency ranking among mask predictions of consecutive frames, encouraging higher consistency for temporally closer frames than for distant ones.
As shown in Fig.~\ref{qualitative_results}, WSRVOS can accurately segment target instances across video frames.

Our proposed WSRVOS framework significantly reduces reliance on costly spatial annotations by utilizing only text supervision, thereby offering a new paradigm for RVOS. Extensive experiments on the A2D-Sentences~\cite{firstRVOS}, JHMDB-Sentences~\cite{firstRVOS}, Refer-YouTube-VOS~\cite{Urvos} and Refer-DAVIS17~\cite{refdavis17} datasets show that our WSROVS significantly outperforms state of the art across all metrics. 
\section{Related work}
\begin{figure*}
\centering
\captionsetup{aboveskip=10pt, belowskip=5pt}
\includegraphics[width=\linewidth]
{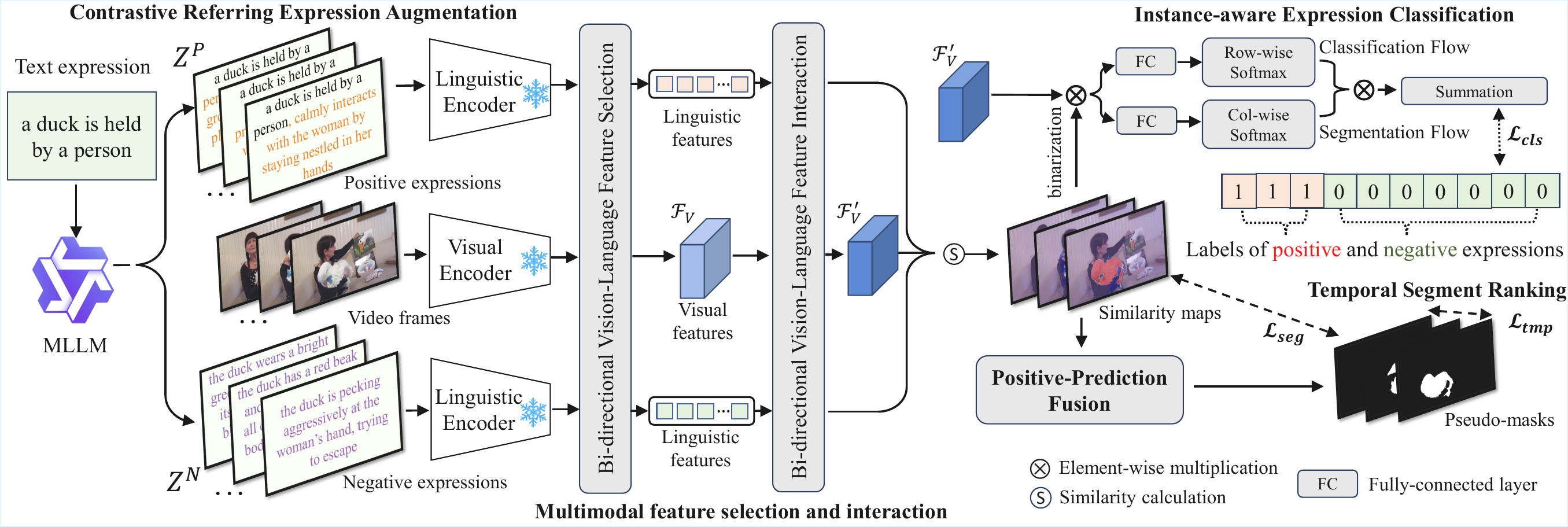}
\caption{Illustration of our proposed WSRVOS. It comprises five main parts: 1) a contrastive referring expression augmentation scheme, leveraging a MLLM to generate positive and negative expressions; 2) a multimodal feature selection and interaction module, facilitating effective multimodal alignment between visual and linguistic features of the given video and generated expressions; 3) an instance-aware expression classification scheme, guiding the model to distinguish between positive and negative expressions; 4) a positive-prediction fusion strategy, generating pseudo-masks as additional supervision signals; 5) finally, a temporal segment ranking constraint, optimizing the overlaps between mask predictions of temporally neighboring frames.}
\label{fig:overview}
\vspace{-5mm}
\end{figure*}
\noindent
\textbf{Referring video object segmentation.}
Existing RVOS methods can be primarily categorized into single-stage and two-stage approaches. Single-stage ones~\cite{RefVOS, CMSA} directly integrate visual and linguistic features to predict masks using a pixel decoder. In contrast, two-stage ones~\cite{AAMN, rethinking} first generate multiple mask candidates and then select the one that most closely matches the text expression to produce the final output.
Recently, the prevalence of query-based Transformer architectures in computer vision \cite{DETR, DeformableDETR} has inspired new innovations in RVOS.
MTTR~\cite{MTTR} adapts the DETR framework~\cite{DETR} to RVOS, while ReferFormer~\cite{ReferFormer} optimizes the Deformable DETR~\cite{DeformableDETR} using fewer instance queries. 
Losh~\cite{losh} segments the target instance under guidance from both long and short text expressions.
ReferDINO \cite{referdino} employs a deformable mask decoder to capture instance-aware dynamics across frames.
OnlineRefer~\cite{onlinerefer} utilizes position prior to improve the accuracy of referring predictions and proposes a cross-frame query propagation module for running the method on the fly.
SAMWISE~\cite{samwise} leverages SAM2~\cite{sam2} as the memory bank to encode long-range context and achieves excellent segmentation results.

\noindent
\textbf{Weakly-supervised referring visual segmentation.}
Several weakly-supervised RVOS methods have been proposed to alleviate the reliance on pixel-level mask annotations \cite{wrvos,ocpg}. Instead, they leverage bounding box or point annotations. For example, WRVOS~\cite{wrvos} requires a pixel-level mask for the frame where the target instance first appears, and bounding box annotations for the remaining frames during training. OCPG~\cite{ocpg} generates pseudo-masks from bounding box or point annotations to train the segmentation model. 
%However, these methods still require manual annotations, which incur considerable annotation efforts and labeling costs for videos. 
In contrast, in the field of weakly-supervised referring image segmentation (RIS), several works have been introduced to achieve accurate segmentation using only text supervision. For instance, TRIS~\cite{TRIS} leverages a text-to-image optimization strategy to locate the target instances.
% using only text supervision during training. 
PCNet~\cite{PCNet} utilizes target-related textual cues from the input expression for progressively localizing the target instance. Inspired by them, we propose a weakly-supervised RVOS framework using solely text supervision. Compared with static images, our WSRVOS is in a more challenging scenario owing to the temporal dynamics and visual redundancy of videos. To cope with them, we propose components such as bi-directional vision-language feature selection module and temporal segment ranking constraint. 

\noindent
\textbf{Multimodal large language models.} 
MLLMs~\cite{Blip2,llava,qwen3,siglip2} have garnered significant attentions due to their impressive performance across a variety of multimodal downstream tasks, such as visual question answering~\cite{llavaonevision} and image captioning~\cite{minigpt}. They learn cross-modal representations by aligning visual and linguistic features through training on a vast corpus of vision-language pairs. Recently, several studies~\cite{villa,visa,onetoken,glus} have explored the use of MLLMs to solve RVOS task. For example, VISA~\cite{visa} employs a frame sampler to select frames most relevant to the text expressions, and then processes visual and linguistic features via MLLMs~\cite{chatunivi} to generate segmentation masks. InstructSeg~\cite{instructseg} leverages MLLMs~\cite{llava} for language-instructed pixel-level reasoning and segmentation.

\section{Method}
\subsection{Problem setting}
The input of our WSRVOS framework consists of a video $\mathcal{V}$ and a text expression $Z^\text{o}$. The objective of WSRVOS is to predict pixel-wise segmentation masks $\mathcal{M}=\left\{m_t\right\}_{t=1}^T$, where $m_t \in \mathbb{R}^{{H} \times {W}}$ is the mask in the $t$-th frame for the target instance referred by $Z^\text{o}$. Different from existing RVOS methods, only $Z^\text{o}$ is available as text supervision. 
The framework of WSRVOS is shown in Fig.~\ref{fig:overview}. It comprises five parts: contrastive referring expression augmentation, multimodal feature selection and interaction, instance-aware expression classification, positive-prediction fusion, and temporal segment ranking.

%During inference, the model can find regions with high response to the input $\mathcal{Q}$ as mask predictions. 

\subsection{Contrastive referring expression augmentation}
\label{sec:generation}
The original expressions in existing RVOS datasets are rather simple and lack fine-grained semantic details regarding visual appearance, action and inter-instance relations. To enrich the training signals, we first use an MLLM (\ie, Qwen3-VL~\cite{qwen3}) to generate positive expressions with richer semantics. For negative expressions, simply selecting expressions from other videos is insufficient due to the low semantic similarity with the given video. Instead, we generate hard negative expressions by using an MLLM to alter the target instance's attributes and actions in $Z^\text{o}$. 
This enables the model to learn more discriminative representations. 

% by changing the target's attributes in $\mathcal{Q}$ or describing other objects via MLLM, enabling the model to learn more discriminative representations.

 % The goal of positive expression generation is to enrich supervision with more  detailed and diverse descriptions of the target object. 

\textbf{Positive expression generation.}  Given $\mathcal{V}$ and $Z^\text{o}$, we first use the following prompt to obtain $P$ descriptions: \textit{`Based on the original text expression $Z^\text{o}$, and considering the video provided, please enrich $P$ positive descriptive sentences focusing on the following aspects: 1. Visual appearance: Describe the instance's color, shape, size, texture, and any distinctive visual features. 2. Action and interaction: Elaborate on the instance's action or interaction with other instances, including any movements, changes, or notable actions it is performing.'}. Since Qwen3-VL~\cite{qwen3} may generate incorrect descriptions, we leverage InternVideo2~\cite{internvideo2} to extract visual feature of $\mathcal{V}$ and linguistic feature of each generated description. Next, we compute their cosine similarity as $cos(\text{IV2}(\mathcal{V}),\text{IV2}({Z}^k))$,
% $\text{InternVideo2}\left( \mathcal{V}, {Z}^k \right)$ between them, 
where ${Z}^k$ represents the $k$-th generated description and $\text{IV2}$ represents InternVideo2. We calculate the confidence score of ${Z}^k$ using:
% \begin{equation}
% c^k = \frac{\text{InternVideo2}\left( \mathcal{V}, {Z}^k \right)}{\text{InternVideo2}\left( \mathcal{V}, {Z^\text{o}} \right)}
% \label{eq:generation}
% \end{equation}
\begin{equation}
c^k = \frac{cos(\text{IV2}(\mathcal{V}),\text{IV2}({Z}^k))}
{cos(\text{IV2}(\mathcal{V}),\text{IV2}(Z^\text{o}))}
\label{eq:generation}
\end{equation}
If $c^k > 0.8$, we regard ${Z}^k$ as an effective description, otherwise, discard it. Finally, we construct the positive expressions by respectively concatenating each description with $Z^\text{o}$ to retain the original information.

% replace it with an empty token.

%an frame-text similarity evaluator. First, we calculate the cosine 
%If the similariti between the sampled frame and original expression as well as generated expression, which are denoted as 
% ${c^P} = \text{Siglip2}\left ( v_s, \mathcal{Q}\right ) / \text{Siglip2}\left ( v_s, \mathcal{Q}\right )$. 

\textbf{Negative expression generation.} Similarly, we utilize Qwen3-VL~\cite{qwen3} to generate $N$ negative expressions through the following prompt: `\textit{Based on the original text expression $Z^\text{o}$, and considering the video provided, please generate $N$ negative descriptive sentences for the given expression focusing on the following aspects: 1. Visual appearance: Describe the instance with incorrect category or attributes (\eg, color, shape, size and texture). 2. Action and interaction: Describe the instance doing a different action or in a different state or with an incorrect spatial relationship.}'  

Through the above scheme, we obtain $P$ positive expressions $\mathcal{Z^P}$ and $N$ negative expressions $\mathcal{Z^N}$ for $\mathcal{V}$. Notably, the MLLM can be employed exclusively for label augmentation in an offline manner for training data and is excluded from the inference process.

\subsection{Multimodal feature selection and interaction}
\label{sec:fusion}
In this stage, we first extract features from $\mathcal{V}$ and generated expressions $\mathcal{Z^P}$ and $\mathcal{Z^N}$. Next, we perform bi-directional vision-language feature selection and interaction. 
%Below we specify the details.
% To facilitate effective cross-modal alignment for subsequent instance localization, we propose a V-L feature selection module that identifies and enhances features most relevant to the target instance. 

\textbf{Visual encoder.} We utilize a pretrained visual encoder to extract visual features from $\mathcal{V}$.
The visual feature of $t$-th frame in the $i$-th encoder block is denoted by 
$v_t^i \in \mathbb{R}^{ N_v \times{C}}$. $N_v$ is the number of visual tokens per frame, and $C$ is the feature dimension, $C = 1024$. The final output of the visual encoder, $\mathcal{F}_V$, is a concatenation of $v_t$ across all frames. 

% $\mathcal{F}_V=\left\{V_t^i\right\}_{t=1}^T$, where 
% $V_t^i \in \mathbb{R}^{ N_v \times{C}}$, $V_t^i &= [{v}_1^i, {v}_2^i, \ldots, {v}_{N_v}^i]$. $D$ is the down-sampling ratio and $N_v = \frac{H}{D} \times \frac{W}{D}$ for usage.

% Each $f_t \in \mathbb{R}^{ \frac{H}{D} \times \frac{W}{D} \times{C_1}}$ in the extracted visual features, $\mathcal{F}_v=\left\{f_t\right\}_{t=1}^T$, corresponds to the $t$-th frame in $\mathcal V$. $D$ is the down-sampling ratio. $f_t$ is later flattened to $\mathbb{R}^{ N_v \times{C_1}}$, where $N_v = \frac{H}{D} \times \frac{W}{D}$ for usage.  

\textbf{Linguistic encoder.} Given positive and negative expressions, $\mathcal{Z^P}$ and $\mathcal{Z^N}$, we extract their linguistic features using a pretrained linguistic encoder. Specifically, the linguistic feature of $k$-th expression in the $i$-th encoder block is denoted by $z_k^i \in \mathbb{R}^{N_l \times C}$, where $N_l$ is the fixed token length (padded with [PAD] tokens if necessary).
% We set the shared feature dimension $C$ for both visual and linguistic features to 1024. 
The final output of the linguistic encoder is denoted as $\mathcal{F}_Z \in \mathbb{R}^{ (P+N) \times{C}}$.

\textbf{Bi-directional vision-language feature selection.} Existing RVOS methods often suffer from inefficient alignment between visual and linguistic features. This is largely due to the temporally dynamic and semantically diverse content that inherently exists in videos. Visual features may include redundant or irrelevant information that does not match the referring expression. In addition, the expressions may also include auxiliary or non-informative words that are unrelated to the video content. This weak alignment may lead to ambiguous or inaccurate localization of the target instance. To address this issue, we select the linguistic and visual features that are highly relevant to each other, enabling fine-grained multimodal alignment.

Specifically, given the visual feature $v_t^i$ and the linguistic feature $z_k^i$ in the $i$-th encoder block, we first compute their cosine similarity and select the top-$K_V$ most relevant visual tokens with respect to $z_k^i$, resulting in ${\widehat{v}}_t^i$. Subsequently, we select the top-$K_Z$ most relevant linguistic tokens with ${\widehat{v}}_t^i$ from $z_k^i$, yielding ${\widehat{z}}_k^i$. These selected visual and linguistic tokens are highly relevant to each other. 
They are restored to their original positions in the token sequence, while the unselected tokens are replaced with zero vectors, so that the refined features remain the same dimension as the original inputs. These refined features are then added to the original features (\ie  $\widecheck{v}_t^{i} = v_t^i + \widehat{v}_t^i$ and $\widecheck{z}_k^{i} = z_k^i + \widehat{z}_k^i$) and are subsequently fed into the next encoder block.

\textbf{Bi-directional vision-language feature interaction.}
We employ a bi-directional attention module to enhance the interaction between final outputs of encoders, \ie, $\mathcal{F}_V$ and $\mathcal{F}_Z$. This enables each modality to fully leverage the information from the other to enrich its own representations. It has two attention streams. The vision-language attention stream treats $\mathcal{F}_Z$ as query while $\mathcal{F}_V$ as key and value in the cross-attention layer, enabling the model to identify which components of the expression are most relevant to the visual content, resulting in $\mathcal{\widetilde{F}}_Z$. In parallel, the language-vision attention stream treats $\mathcal{F}_V$ as query to highlight visual regions that are most relevant to the referring expression, generating refined visual features $\mathcal{\widetilde{F}}_V$. Next, we employ the feed-forward network (FFN) with skip-connection and layer normalization to respectively obtain ${\mathcal{F}'_V} = \text{LayerNorm}(\text{FFN}(\mathcal{\widetilde{F}}_V) + {\mathcal{F}_V})$
and ${\mathcal{F}'_Z} = \text{LayerNorm}(\text{FFN}(\mathcal{\widetilde{F}}_Z) + {\mathcal{F}_Z})$. 
% Finally, we apply $L_2$ normalization along the channel dimension for both modalities.

% \begin{figure}[htbp]
% \centering
% \includegraphics[width=\linewidth]
% {figures/Figure3.png}
% \caption{The temporal segment ranking constraint.}
% \label{fig:tsr}
% \vspace{-5mm}
% \end{figure}

\subsection{Instance-aware expression classification}
\label{sec:cls}
% Each $f_t \in \mathbb{R}^{ \frac{H}{D} \times \frac{W}{D} \times{C_1}}$ in the extracted visual features, $\mathcal{F}_v=\left\{f_t\right\}_{t=1}^T$, corresponds to the $t$-th frame in $\mathcal V$.
% Although we can follow~\cite{TRIS} to directly perform classification based on the proposals to 
After obtaining ${\mathcal{F}'_V}=\left\{v_t\right\}_{t=1}^T$ and ${\mathcal{F}'_Z}$, the next step is to compute their similarity maps and train the model to distinguish between positive and negative expressions. Specifically, we first compute the vision-language similarity maps $S=\left\{{s_t}\right\}_{t=1}^{T}$ by performing matrix multiplication between the visual feature of $t$-th video frame (\ie, $v_t \in \mathbb{R}^{ N_v \times{C}}$) and the transpose of ${\mathcal{F}'_Z}$, where $s_t \in \mathbb{R}^{N_v \times (P+N)}$ since we have $P$ positive expressions and $N$ negative expressions. For notational simplicity, we omit the frame index $t$ in the following descriptions unless otherwise specified (\eg, $s_t$ is denoted as $s$, $v_t$ is denoted as $v$). 

Accordingly, each element ${s}_{j}^{k}$ in $s$ indicates the similarity between the \textit{k}-th referring expression and the \textit{j}-th patch in the frame. A higher value of ${s}_{j}^{k}$ indicates a higher likelihood that the patch belongs to the instance referred by the expression, which can be used to locate the target instance.
%while a lower value of $\mathbf{s}_{t,j}^{k}$ indicates that the patch is more likely corresponds to irrelevant background or distractors. 
We apply a sigmoid activation and threshold ($\theta = 0.4$) filtering on {the vector} $s^k\in \mathbb{R}^{N_v}$ to obtain a binary mask $m^k \in \mathbb{R}^{N_v}$, which indicates the patches in the frame that exhibit high similarities to the \textit{k}-th referring expression, \ie, a proposal for the \textit{k}-th referring expression. Next, we compute the feature representation of the proposal $r^k \in \mathbb{R}^{C}$ by averaging the visual feature $v$ 
over the foreground patches (1-valued) identified by $m^k$: 
% \begin{equation}
% P_t^k = \frac{1}{\sum_{j=1}^{N_v} m_t^{j,k}} \sum_{j=1}^{N_v} m_t^{j,k} \cdot f_t^{j,k}
$r^k = \frac{1}{\sum_{j=1}^{N_v} m_{j}^{k}} \sum_{j=1}^{N_v} m_{j}^{k} \cdot v_{j}$,
% \end{equation}
% where $f_t^{j,k}$ denotes the visual feature of the $j$-th patch in the $t$-th frame, enhanced by the interaction with the \textit{k}-th expression. 
where $v_{j}$ denotes the visual feature of the $j$-th patch in the frame. 
% It is possible that  negative expressions may correspond to no activated patches, we assign zero vectors as their feature representations. 
In case when certain expression does not correspond to any activated patches, we assign a zero vector as its feature representation.
In this way, we can obtain in total $P+N$ instance-aware proposal features, which are  concatenated into $R \in \mathbb{R}^{(P+N) \times C}$.
%$P$ positive expressions and $N$ negative expressions. 
 
% the average of visual features of patches that are highly relevant to the \textit{k}-th expression.

%However, not all positive expressions contribute equally to the localization of the target instance. 
Recalling that positive expressions are augmented from different perspectives, therefore they contribute unequally to the localization of the target instance.  
%Some may describe discriminative visual attributes, while others may be vague or only partially aligned with the target instance. 
On the other hand, negative expressions are incorrect descriptions for the target instance, but still share partial similarities with the instance (\eg referring to the same instance category but differing in attributes). %Therefore, simply treating all expressions and their associated proposals with equal contribution in the model may introduce noise and degrade performance. 
Motivated by this, we perform \emph{proposal aggregation and expression matching} based on multiple instance learning (MIL)~\cite{WSDDN,MIL}, which enables the model to assess each proposal's  contribution to the matching score between the video frame and certain expression.
% , emphasizing more informative proposals while suppressing less informative ones. 
Below we specify the details.

\textbf{Proposal aggregation and expression matching.}
\label{sec:mil}
We pass $R$ through a MIL structure consisting of two parallel fully-connected layers, which can be regarded as a classification flow and a segmentation flow. Their parameters are denoted by $W^{cls}\in \mathbb{R}^ {C \times (P+N)}$ and $W^{smt} \in \mathbb{R}^ {C \times (P+N)}$, respectively. These two flows output two matrices of scores $U^{cls} = RW^{cls}$ and $U^{smt} = RW^{smt}$, where $U^{cls}, U^{smt} \in \mathbb{R}^{(P+N) \times (P+N)}$. The rows of $U^{cls}, U^{smt}$ correspond to the $P+N$ proposals while the columns correspond to the $P+N$ expressions. We apply a row-wise softmax to $U^{cls}$ and a column-wise softmax to $U^{smt}$. Hence, the classification flow is designed to predict which expression is associated with each proposal, whereas the segmentation flow 
selects which proposals are likely to contain informative frame fragments (relevant to certain expression).
%predicts which proposals are more likely to certain expression. 
In pure vision task, $W^{cls}$ is typically set to be learnable to perform proposal classification. However, in our multimodal task, it aims to determine the expression to which each proposal corresponds. Therefore, we can directly set $W^{cls}$ using the transpose of linguistic features of all expressions (\ie ${\mathcal{F}_Z}$).
% , enabling the model to compute proposal-expression matching scores $U^{cls}$.
In contrast, $W^{smt}$ is learnable.

Next, we combine the proposal scores from two flows, $U^{cls}$ and $U^{smt}$, through element-wise multiplication to obtain $U^{fuse} \in \mathbb{R}^ {(P+N) \times (P+N)}$. Each element $u_{n}^k$ in $U^{fuse}$ signifies the matching score between the $n$-th proposal and the $k$-th expression. 
%We then transform these proposal-expression  matching scores $U^{fuse}$ into frame-expression matching scores $y \in \mathbb{R}^{(P+N)}$ by summing over the row dimension of $U^{fuse}$, 
%Each column of $U^{fuse}$ 
In each column of $U^{fuse}$, we can aggregate multiple proposal-expression scores into one frame-expression score by summing up all the row values inside:
\begin{equation}
y^k = \sum_{n=1}^{P+N} u_{n}^k
\end{equation}
We apply the binary cross-entropy loss to supervise the classification of positive and negative expressions:
% The classification loss is defined using the binary cross-entropy loss:
\begin{equation}
\begin{aligned}
\mathcal{L}_{\text{cls}}= 
- \frac{1}{P+N} \, & \sum_{k=1}^{P+N} \Bigg[
g^{k} \log \left( \frac{1}{1 + e^{-y^k}} \right) \\
& \quad + (1 - g^{k}) \log \left( \frac{e^{-y^k}}{1 + e^{-y^k}} \right)
\Bigg]
\end{aligned}
\end{equation}
\noindent
where $g^k$ denotes the ground truth indicating whether the \textit{k}-th expression for the frame is a positive one ($g^k =  1$) or not ($g^k = 0$).

% Each entry $y_t^k$ reflects the overall confidence that the $k$-th expression corresponds to the visual content of the $t$-th frame, aggregated over all available proposals.

% \begin{equation}
% \mathcal{L}_{mil}(\widetilde{y}, z)=-\frac{1}{P+N}  \sum_{t=1}^{T}\sum_{k=1}^{P+N}  z_t^{k} \log \left(\frac{1}{1+e^{-\widetilde{y_t^k}}}\right)  +\left(1-z_t^{k}\right) \log \left(\frac{e^{-\widetilde{y_t^k}}}{1+e^{-\widetilde{y_t^k}}}\right)
% \end{equation}
% where $z_t^k $ is the same binary ground truth label used in Sec.~\ref{sec:cls}. Different from  $y$ in Sec.~\ref{sec:cls}, $\widetilde{y}$ represents the probability that an instance-level proposal is referred by a specific expression, rather than an image-level frame. The reason is that $\widetilde{y}$ is computed from visual features likely to be associated with regions containing the referred instance, rather than being computed from all patch features like ${y}$. Therefore, $\mathcal L_{mil}$ can align instance-level visual features with linguistic features by optimizing $\widetilde{y}$.
\renewcommand{\arraystretch}{1.1}
\begin{table*}
\centering
\captionsetup{aboveskip=2pt, belowskip=0pt}
\resizebox{\textwidth}{!}{%
\begin{tabular}{ccccccccccccc}
\hline
\multirow{2}{*}{\textbf{Supervision}} & \multirow{2}{*}{\textbf{Method}} & \multirow{2}{*}{\textbf{Backbone}} 
& \multicolumn{2}{c}{\textbf{A2D-Sentences}} 
& \multicolumn{2}{c}{\textbf{JHMDB-Sentences}} 
& \multicolumn{3}{c}{\textbf{Ref-YouTube-VOS}} 
& \multicolumn{3}{c}{\textbf{Ref-DAVIS17}} \\
\cline{4-13}
& & & O-IoU & M-IoU & O-IoU & M-IoU & $\mathcal{J}$ & $\mathcal{F}$ & $\mathcal{J}$\&$\mathcal{F}$ & $\mathcal{J}$ & $\mathcal{F}$ & $\mathcal{J}$\&$\mathcal{F}$ \\
\hline
\multirow{6}{*}{Text + Mask} 
& MTTR~\cite{MTTR} & Video-Swin-T & 70.2 & 61.8 & 67.4 & 67.9 & 54.0 & 56.6 & 55.3 & 53.2 & 55.9 & 54.6\\
& ReferFormer~\cite{ReferFormer} & Video-Swin-T & 77.6 & 69.6 & 71.9 & 71.0 & 58.0 & 60.9 & 59.4 & 56.5 & 62.7 & 59.6\\
& SOC~\cite{soc} & Video-Swin-T & 78.3 & 70.6 & 72.7 & 71.6 & 61.1 & 63.7 & 62.4 & 60.2 & 66.7 & 63.5\\
& DsHmp~\cite{dshmp} & Video-Swin-T & 79.0 & 71.3 & 73.1 & 72.1 & 61.8 & 65.4 & 63.6 & 60.8 & 67.2 & 64.0\\
& LoSh~\cite{losh} & Video-Swin-T & 79.3 & 71.6 & 73.6 & 72.7 & 62.0 & 65.4 & 63.7 & 60.1 & 65.7 & 62.9\\
& ReferDINO~\cite{referdino} & Video-Swin-T & \cellcolor{gray!15}80.2 & \cellcolor{gray!15}72.3 & \cellcolor{gray!15}74.2 & \cellcolor{gray!15}73.1 & \cellcolor{gray!15}65.5 & \cellcolor{gray!15}69.6 & \cellcolor{gray!15}67.5 & \cellcolor{gray!15}62.9 & \cellcolor{gray!15}70.7 & \cellcolor{gray!15}66.7\\
\hline
\multirow{5}{*}{Text + Box} 
& BoxInst~\cite{boxinst} & ViT-B & 60.1 & 46.4 & 62.9 & 60.3 & 43.8 & 49.4 & 46.6 & 41.4 & 48.3 & 44.9\\
& BoxLevelSet~\cite{boxlevelset} & ViT-B & 56.2 & 51.2 & 64.6 & 64.6 & 45.4 & 50.7 & 48.0 & 43.5 & 50.1 & 46.8\\
& BoxVOS~\cite{boxvos} & Video-Swin-T & 59.6 & 53.2 & 65.0 & 64.8 & 46.1 & 51.7 & 48.9 & 44.3 & 52.0 & 48.2\\
& WRVOS~\cite{wrvos} & Video-Swin-T & \cellcolor{gray!15}66.3 & 53.9 & 63.2 & 62.7 & 48.9 & 51.4 & 50.2 & 45.6 & 51.1 & 48.3\\
& OCPG~\cite{ocpg} & Video-Swin-T & 65.0 & \cellcolor{gray!15}57.8 & \cellcolor{gray!15}68.9 & \cellcolor{gray!15}68.6 & \cellcolor{gray!15}50.4 & \cellcolor{gray!15}55.1 & \cellcolor{gray!15}52.8 & \cellcolor{gray!15}48.1 & \cellcolor{gray!15}53.5 & \cellcolor{gray!15}50.8\\
\hline
\multirow{4}{*}{Text + Point} 
& PPT*~\cite{ppt} & Video-Swin-T & 36.8 & 33.3 & 47.9 & 47.0 & 19.4 & 21.2 & 20.8 & 19.7 & 21.4 & 20.6 \\
& PointVOS~\cite{pointvos} & Video-Swin-T & 43.0 & 39.2 & 54.5 & 52.6 & 25.1 & 32.5 & 29.1 & 24.8 & 27.4 & 26.1\\
& OCPG~\cite{ocpg} & Video-Swin-T
& \cellcolor{gray!15}{48.8}
& \cellcolor{gray!15}{45.0}
& \cellcolor{gray!15}{58.3}
& \cellcolor{gray!15}{57.9}
& \cellcolor{gray!15}{33.0}
& \cellcolor{gray!15}{40.8}
& \cellcolor{gray!15}{36.9}
& \cellcolor{gray!15}{31.8}
& \cellcolor{gray!15}{39.4}
& \cellcolor{gray!15}{35.6} \\
& SEVOS~\cite{sevos} & Video-Swin-T & 46.5 & 42.8 & 57.5 & 56.7 & 29.6 & 36.5 & 33.0 & 29.0 & 32.8 & 30.9 \\
\hline
\multirow{5}{*}{Text} 
% & CLIP-MIL~\cite{clipmil} & Video-Swin-T & 31.7 & 30.7 & 44.1 & 45.2 & & & & & & \\
% & ReferredAdapter~\cite{referredadapter} & Video-Swin-T & 29.7 & 32.7 & 46.7 & 47.3 & & & & & & \\
& TRIS*~\cite{TRIS} & Video-Swin-T & 34.2 & 30.8 & 47.2 & 46.5 & 16.8 & 20.1 & 18.5 & 15.9 & 17.0 & 16.5\\
& PCNet*~\cite{PCNet} & Video-Swin-T & 38.8 & 34.6 & 49.7 & 48.9 & 17.9 & 21.6 & 19.7 & 19.0 & 20.1 & 19.6\\
& DViN*~\cite{dvin} & Video-Swin-T & 41.1 & 38.3 & 51.8 & 50.9 & 26.1 & 32.7 & 29.4 & 25.3 & 30.9 & 28.1\\
& WSRVOS~(Ours) & Video-Swin-T
& \cellcolor{gray!15}{\textbf{48.2}}
& \cellcolor{gray!15}{\textbf{44.0}}
& \cellcolor{gray!15}{\textbf{58.7}}
& \cellcolor{gray!15}{\textbf{58.1}}
& \cellcolor{gray!15}{\textbf{33.5}}
& \cellcolor{gray!15}{\textbf{39.4}}
& \cellcolor{gray!15}{\textbf{36.5}}
& \cellcolor{gray!15}{\textbf{32.6}}
& \cellcolor{gray!15}{\textbf{38.2}}
& \cellcolor{gray!15}{\textbf{35.4}} \\
& WSRVOS~(Ours) & Video-Swin-B
& \cellcolor{gray!15}{\textbf{50.6}}
& \cellcolor{gray!15}{\textbf{46.2}}
& \cellcolor{gray!15}{\textbf{60.1}}
& \cellcolor{gray!15}{\textbf{59.2}}
& \cellcolor{gray!15}{\textbf{37.2}}
& \cellcolor{gray!15}{\textbf{42.0}}
& \cellcolor{gray!15}{\textbf{39.6}}
& \cellcolor{gray!15}{\textbf{35.2}}
& \cellcolor{gray!15}{\textbf{39.8}}
& \cellcolor{gray!15}{\textbf{37.5}} \\
\hline
\end{tabular}%
}
\caption{Comparison to state of the art under different supervision types (Text + Mask, Text + Box, Text + Point, Text). 
O-IoU and M-IoU represent Overall IoU and Mean IoU. 
$\mathcal{J}$ and $\mathcal{F}$ denote region similarity and contour accuracy respectively.
$\mathcal{J}$\&$\mathcal{F}$ is the average of $\mathcal{J}$ and $\mathcal{F}$. 
* indicates methods adapted from weakly-supervised RIS for RVOS task. We mark the best performance for each supervision type in \colorbox{gray!15}{shadow}.}
\label{tab:SOTAcomparison_combined}
\vspace{-2mm}
\end{table*}

% The proposed $\mathcal L_{cls}$ and $\mathcal L_{mil}$ start from the perspective of image-level and instance-level feature alignment respectively, enabling the model to locate referred target instances using only text supervision.
%Previously, enriched positive and negative expressions are used to improve the model's image-level and instance-level multimodal feature alignment capabilities. 
%use various positive expressions to generate comprehensive and accurate masks as pixel-wise supervision, directly improving the segmentation accuracy and robustness of our model.

\subsection{Positive-prediction fusion}
\label{sec:seg}
In this section, we focus on fusing the predictions from enriched positive expressions to generate pseudo-masks, enabling effective supervision to improve segmentation.

% To this end, we propose a positive-prediction fusion strategy, that can comprehensively consider high-response regions corresponding to multiple input $\mathcal{Q^P}$ to generate  a high-quality fused similarity map to supervise our model. 
We propose a positive-prediction fusion strategy, which aggregates the likely regions (high vision-language similarities) corresponding to multiple positive expressions to construct a pseudo-mask for supervision. Specifically, we re-weight each similarity map $s^k$ using the confidence score $c^k$ obtained in the positive expression generation stage (\ie, Eq.~(\ref{eq:generation})). The fusion process is written as following:

\begin{equation}
s^{final} = \frac{\sum_{k=1}^{P} s^{k} \times c^{k}}{\sum_{k=1}^{P} c^{k}} 
\end{equation}
where $s^{k}$ denotes the similarity map between the \textit{k}-th positive expression and the frame. After obtaining $s^{final}$, 
% we can obtain the final pixel-wise pseudo-label, $z_t^{mask} \in \mathbb{R}^{N_v}$, by performing sigmoid operation and threshold filtering. The segmentation loss consisting of a mask focal loss and a DICE loss can be formulated as:
we apply the sigmoid operation and threshold filtering on it to generate the binary pseudo-mask $m \in \mathbb{R}^{N_v}$.
The segmentation loss consists of a Focal loss and a DICE loss applied between $m$ and each $s^{k}$:
\begin{equation}
\label{equ5}
\begin{aligned}
\mathcal{L}_{\text{seg}} =
\frac{1}{P} \, &  \sum_{k=1}^{P} \, 
\text{Focal} \left( s^k, m \right) +  \text{DICE} \left( s^k, m \right)
\end{aligned}
\end{equation}
\noindent
% where $\lambda_{1}$ is a hyper-parameter.

%We further propose a novel segmentation temporal ranking constraint to integrate the temporal factor into the optimization. 
%we observe that object motion is either tends to either increase (moving objects) or remain stable (static objects); therefore object spatial layouts in closer frames are expected to be more consistent than those in farther ones.
% Specifically, our observation is that objects normally move smoothly over frames for a short period of time, therefore their spatial layouts in temporally closer frames are expected to be more consistent than those in farther ones. 
% This means, given $n+1$ neighboring frames with pseudo-masks denoted as $g_t, g_{t+\delta},g_{t+2\delta},\dots, g_{t+n\delta}$, their pairwise IoU value should ideally satisfy the ranking: $\text{IoU}(g_t,g_{t+\delta})\ge \text{IoU}(g_t,g_{t+2\delta})\ge \cdots\ge \text{IoU}(g_t,g_{t+n\delta})$, where $\delta$ denotes a predefined temporal interval. In practice, we set $n=3$ and optimize the following constraint:

% $\mathcal{L}_{\text{seg}}$ optimizes each frame's vision-language similarity map towards its pseudo-mask.

\subsection{Temporal segment ranking}
\label{sec:temporal}
Once we obtain predicted masks for the target instance in a short video, because of the instance's motion effect, overlap between instance masks either gets decreased with an increase of time or remains stable if the motion is insignificant.  
% This means that, for illustration, we consider four temporally neighboring frames with generated pseudo-masks $m_1, m_2, m_3, m_4$. Their pairwise IoU value should normally satisfy the ranking: $\text{IoU}(m_1,m_{2})\ge \text{IoU}(m_1,m_{3})\ge \text{IoU}(m_1,m_{4})$.
% , where $\delta$ denotes a predefined temporal interval. 
% We first fix a predefined set of temporal intervals:
% $\mathcal{D}$ = \{3, 6, 9\}
% where each element in $\mathcal{D}$ denotes the temporal distance between a reference frame and a comparison frame. For a video containing $T$ frames (let $m_t$ represent the pseudo-mask of the $t$-th frame), we require the following constraint: for any reference frame $t$, and for any two intervals $d_1, d_2 \in \mathcal{D}$ satisfying $d_1 < d_2$, the IoU between $m_t$ and the pseudo-mask of the frame at interval $d_1$ (i.e., $m_{t+d_1}$) should be no smaller than the IoU between $m_t$ and the pseudo-mask of the frame at interval $d_2$ (i.e., $m_{t+d_2}$).
This means that, for illustration, we consider  $T$ temporally sequential frames in a short video. The generated pseudo-masks are denoted as $\mathcal{M}=\left\{m_t\right\}_{t=1}^T$. Their pairwise IoU value should normally satisfy the temporal ranking: $\text{IoU}(m_t, m_l) \ge \text{IoU}(m_t, m_n)$, where $1\le t< l< n\le T$. In practice, we optimize the following constraint:
\begin{equation}
\label{equ7}
\mathcal{L}_{\text{tmp}} \;=\; 
\sum_{1 \le t < l< n \le T}
f\!\big( \text{IoU}(m_t,m_n) - \text{IoU}(m_t,m_l) - \varepsilon \big),
\end{equation}
% \begin{equation}
% \label{equ7}
% \mathcal{L}_{\text{tmp}} = \sum_{t=1}^{T} \sum_{\substack{d_1 < d_2 \\ d_1, d_2 \in \mathcal{D}}} f\left(  \left( \text{IoU}(m_t, m_{t+d_2}) -  \text{IoU}(m_t, m_{t+d_1}) - \varepsilon  \right) \right)
% \end{equation}
% \begin{equation}
% \label{equ7}
% \mathcal{L}_{\text{tmp}}
% = \sum_{1 \le j < k \le 3}
% f\Bigl(
% \varepsilon - \bigl(\mathrm{IoU}(m_1,m_{1+j}) - \mathrm{IoU}(m_1,m_{1+k})\bigr)
% \Bigr)
% \end{equation}
where $f(\cdot )$ is the \emph{ReLU} function and $\varepsilon$ is a hyperparameter that controls the constraint strength. A larger $\varepsilon$ indicates a greater tolerance for deviations from the ground truth ranking, while a smaller $\varepsilon$ enforces the ranking more strictly.

\subsection{Model training and inference}
\label{sec:training-inference}
Based on the above stages, our WSRVOS can be trained in an end-to-end manner.  The overall training objective is defined as: 
\begin{equation}
\label{equ8}
  \mathcal{L}
  \;=\;
  \mathcal{L}_{\text{cls}}
  \;+\;
  \lambda_1\,\mathcal{L}_{\text{seg}}
  \;+\;
  \lambda_2\,\mathcal{L}_{\text{tmp}}\,,
\end{equation}
where $\lambda_{1}$ and $\lambda_{2}$ are hyper-parameters. 

During inference, no additional positive or negative expressions are required. We directly extract the visual feature of each video frame and the linguistic feature of the input referring expression. Next, we compute their similarity map $s$, which is then binarized using the threshold ($\theta=0.4$) to obtain the instance segmentation mask. 

% We first compute the similarity map $s$ between the input referring expression ${Z^{\text{o}}}$ and the visual feature of each video frame. This map is binarized using the threshold ($\theta$) to obtain the instance segmentation mask. 

% \subsection{Model training and inference}
% \label{sec:final}
% Based on the above stages, our WSRVOS can be trained in an end-to-end manner. The overall training objective is defined as: 
% \begin{equation}\label{equ8}
% \mathcal {L} = \mathcal L_{cls} + \lambda_{2}  \mathcal L_{seg}
% \end{equation}
% where $\lambda_{2}$ is a hyper-parameter.

% During inference,  we can directly upsample and binarize the similarity map $s_t$ between the input referring expression ${Z^\text{o}}$ and video frames to generate mask predictions. 

% During inference, we first compute the similarity map $s$ between the input referring expression ${Z^{\text{o}}}$ and the visual feature of each video frame. This map is binarized using the threshold ($\theta$) to obtain the instance segmentation mask. 
%using a threshold to identify the expression-relevant patches. Patches with similarity scores above the threshold are set to 1, while the remaining patches are set to 0 to generate the predicted segmentation mask.

% $\mathcal{M}=\left\{m_t\right\}_{t=1}^T$, where $m_t \in \mathbb{R}^{{H} \times {W}}$.
\section{Experiments}
\label{sec:exp}
\subsection{Datasets and evaluation metrics}
We conduct experiments on four widely-used RVOS benchmarks: A2D-Sentences~\cite{firstRVOS}, JHMDB-Sentences~\cite{firstRVOS},  Refer-YouTube-VOS~\cite{Urvos} and Refer-DAVIS17~\cite{refdavis17}. The dataset details are provided in the supplementary material.
% A2D-Sentences contains 3,754 videos, split into 3,017 for training and 737 for testing. 
% % Each video includes annotations for three or five frames, providing pixel-wise segmentation masks across various target instances and accompanied by 6,655 text expressions.  Each referring expression refers to a distinct target instance within the annotated frames of a video.
% Each video is annotated with three or five frames, providing pixel-wise segmentation masks for various target instances. In total, the dataset includes 6,655 referring expressions, each of which corresponds to a instance within the annotated frames.
% Refer-YouTube-VOS~\cite{Urvos}, the largest RVOS dataset to date, comprises 3,978 videos annotated with 15,009 referring expressions. Pixel-wise segmentation masks are provided for every fifth frame in these videos. We train our model on the official training split and report performance on the validation split. 
% Furthermore, following~\cite{ReferFormer, SgMg}, we evaluate the models trained on A2D-Sentences and Refer-YouTube-VOS on JHMDB-Sentences and Refer-DAVIS17 without finetuning, respectively. These two datasets extend the original vision-only datasets, JHMDB~\cite{visualjhmdb} and DAVIS17~\cite{davis17}, by incorporating rich text annotations. Specifically, JHMDB-Sentences is comprised of 928 videos, each paired with a referring expression. In contrast, Refer-DAVIS17 includes 90 videos and is annotated with a total of 1,544 referring expressions.

Following~\cite{firstRVOS}, we utilize Overall IoU and Mean IoU to evaluate our model on A2D-Sentences and JHMDB-Sentences datasets. 
% Overall IoU measures the intersection-over-union between all predictions and ground-truths across the entire test set. Mean IoU calculates the average IoU for each annotated instance across all the frames in the video. 
For Refer-YouTube-VOS and Refer-DAVIS17, we follow~\cite{holistic} to leverage the region similarity ($\mathcal{J}$), contour accuracy ($\mathcal{F}$) and their mean ($\mathcal{J}$\&$\mathcal{F}$) as the evaluation metrics. 
% Note that we do not report contour accuracy ($\mathcal{F}$) in our evaluation. 
%This is because our WSRVOS currently lacks strong boundary prediction capabilities due to the absence of pixel-level supervision during training.

%whereas $\mathcal{F}$ computes the F1-score based on the precision and recall of the contour delineating the predicted target instance.

% Following~\cite{firstRVOS, MTTR}, we adopt  Overall IoU, Mean IoU and mAP to evaluate our model on A2D-Sentences and JHMDB-Sentences.  Overall IoU calculates the ratio of the intersection and union of all predicted and annotated segment pixels in the test set. Mean IoU calculates the average of IOU results for each instance over all frames in the validation set. mAP computes the average of mAPs calculated under 10 IoU thresholds between 0.5 and 0.95 over testing samples. For Refer-YouTube-VOS and Refer-DAVIS17, we follow~\cite{ReferFormer} to use region similarity ($\mathcal{J}$),  contour accuracy ($\mathcal{F}$)  and the average of them ($\mathcal{J} \& \mathcal{F}$). The calculation of $\mathcal{J}$ is the same as IoU between prediction and annotation. $\mathcal{F}$ is the F1-score calculated by the precision and recall of the  contour of the predicted mask for the target instance. 

\subsection{Implementation details}

% frames
% hidden dimensions 
We follow~\cite{soc} to utilize the pre-trained Video-Swin-Tiny~\cite{videoswint} and RoBERTa~\cite{roberta} as our visual and linguistic encoders by default, respectively. 
During training, the parameters of the encoders are frozen. 
The number of positive and negative expressions, $P$ and $N$, is 6 and 48, respectively. 
The vision-language feature selection module is integrated after the 3rd, 6th, 9th, 12th layers of encoders.  
The number of selected visual and linguistic tokens $K_V$ and $K_Z$ is both set to 10. 
%We set the feature dimension $C$ for visual and linguistic features to 1024. 
% The threshold $\theta$ used to obtain binary masks is set to 0.4. 
The parameters $\varepsilon$ in Eq.~(\ref{equ7}) is set to 0.1. The loss weights $\lambda_{1}$ and $\lambda_{2}$ in Eq.~(\ref{equ8}) are set to 2 and 1. Above hyper-parameters are determined based on the validation set of Refer-YouTube-VOS.
During training, we follow~\cite{ReferFormer} to randomly sample $T=4$ frames per video at 10-frame intervals. 
% All frames are resized to $360 \times 640$. 
% Data augmentation includes horizontal flipping with a probability of 0.5, along with corresponding updates to direction-related terms in the text expressions to maintain consistency.
We optimize the model using the AdamW optimizer with an initial learning rate of $1e^{-4}$ and a weight decay of $5e^{-4}$. The number of training epochs is 50. All experiments are conducted using four NVIDIA Tesla A40 GPUs.

% \begin{figure}
% \begin{center}

% \includegraphics[width=\linewidth]
% {figures/Figure4.png}

% \end{center}
%    \caption{Qualitative results across video frames.}
% \label{qualitative_results}
% \end{figure}

\subsection{Comparison with state of the art}
% \hl{Since we are the first weakly-supervised RVOS work only utilizing text supervision, we adapt the recent weakly-supervised RIS methods (TRIS~\cite{TRIS}, PCNet~\cite{PCNet}, DViN~\cite{dvin}) into RVOS for comparison. We also report the performance of several weakly-supervised RVOS methods using bounding box or point supervision (WRVOS~\cite{wrvos}, OCPG~\cite{ocpg}) and some fully-supervised RVOS models for references (SOC~\cite{soc}, LoSh~\cite{losh}, ReferDINO~\cite{referdino}). }

We compare the proposed WSRVOS with six fully-supervised RVOS methods (MTTR~\cite{MTTR}, ReferFormer~\cite{ReferFormer}, SOC~\cite{soc}, DsHmp~\cite{dshmp}, LoSh~\cite{losh}, ReferDINO~\cite{referdino}) and four recent weakly-supervised RVOS methods (WRVOS~\cite{wrvos}, PointVOS~\cite{pointvos}, OCPG~\cite{ocpg} and SEVOS~\cite{sevos}). Note that these works rely on additional bounding box or point annotations, whereas our WSRVOS only uses text supervision. 
% Notably, OCPG~\cite{ocpg} supports both point-supervised and box-supervised settings. 
% In the point-supervised setting like ~\cite{pointvis}, point annotations are synthesized by randomly sampling points given the ground truth mask in each frame. 
For a more comprehensive evaluation, we also adapt four weakly-supervised RIS methods (PPT~\cite{ppt}, TRIS~\cite{TRIS}, PCNet~\cite{PCNet}, and DViN~\cite{dvin}) into RVOS for performance comparison.

As shown in Tab.~\ref{tab:SOTAcomparison_combined}, our WSRVOS achieves excellent performance across all metrics.  
For instance, utilizing only text supervision, WSRVOS achieves +7.1\% improvement in Overall IoU and +5.7\% in Mean IoU compared with DViN on A2D-Sentences. Moreover, WSRVOS significantly outperforms DViN: +7.1\% $\mathcal{J}$\&$\mathcal{F}$ on Refer-YouTube-VOS and +7.3\% $\mathcal{J}$\&$\mathcal{F}$ on Refer-DAVIS17, highlighting its superior segmentation performance and cross-dataset generalizability. Note that WSRVOS even achieves comparable performance with the best point-supervised method OCPG~\cite{ocpg}, \eg, WSRVOS achieves +0.4\% Overall IoU and +0.2\% Mean IoU on JHMDB-Sentences, +0.5\% $\mathcal{F}$ on Refer-YouTube-VOS and +0.8\% $\mathcal{F}$ on Refer-DAVIS17. The point-supervised methods still require manually labeling points on multiple frames for each video, while we do not; for example, SEVOS~\cite{sevos} annotate each frame in the video. Furthermore, when equipped with a larger visual encoder (\ie Video-Swin-B), WSRVOS achieves further performance gains.

We also evaluate the RVOS performance on the A2D-Sentences by directly inquiring the pre-trained MLLMs (\ie, Video-LLaVA~\cite{videollava}, VideoLLaMA3~\cite{videollama}, Qwen3-VL~\cite{qwen3}) with the given video and the prompt: ``You need to perform referring video object segmentation on the given video according to the provided text expression: [referring expression]". As shown in Tab.~\ref{ablation:MLLM}, directly using MLLMs to perform RVOS results in limited performance, indicating the necessity of task-specific fine-tuning for RVOS.

\begin{table}[!t]
\centering
\captionsetup{font=small}
\captionsetup{aboveskip=2pt, belowskip=0pt}
\begin{minipage}[t]{0.48\linewidth}
  \vspace{0pt}
  \centering
  \footnotesize
  \renewcommand{\arraystretch}{1.1}
  \setlength{\tabcolsep}{0.8mm}
  \resizebox{\linewidth}{!}{%
    \begin{tabular}{@{}c|cc@{}}
      \hline
      \textbf{Method} & \textbf{O-IoU} & \textbf{M-IoU} \\
      \hline
      Video-LLaVA-7B~\cite{videollava} & 15.8 & 11.1 \\
      VideoLLaMA3-7B~\cite{videollama} & 18.2 & 14.5 \\
      Qwen3-VL-8B~\cite{qwen3}         & 24.5 & 19.3 \\
      WSRVOS                            & \textbf{48.2} & \textbf{44.0} \\
      \hline
    \end{tabular}
  }% end resizebox
  \caption{Performance of pre-trained MLLMs in RVOS on A2D-sentences.}
  \label{ablation:MLLM}
\end{minipage}
\hfill
\begin{minipage}[t]{0.48\linewidth}
  \vspace{0pt}
  \centering
  \footnotesize
  \renewcommand{\arraystretch}{1.1}
  \setlength{\tabcolsep}{0.8mm}
  \resizebox{\linewidth}{!}{%
    \begin{tabular}{@{}c|cc@{}}
      \hline
      \textbf{Method} & \textbf{Parms(M)} & \textbf{Infer(FPS)} \\
      \hline
      ReferDINO~\cite{referdino} & 128 & 31 \\
      OCPG~\cite{ocpg}           & 58  & 37 \\
      DViN~\cite{dvin}           & 54  & 53 \\
      WSRVOS                     & \textbf{31} & \textbf{58} \\
      \hline
    \end{tabular}
  }% end resizebox
  \caption{Efficiency comparison between different methods.}
  \label{tab:efficiency}
\end{minipage}
% \vspace{-2mm}
\end{table}

 % \\

% \begin{table}[!htbp]
% \centering
% \begingroup
% \setlength{\tabcolsep}{0.7mm}
% \setlength{\textfloatsep}{8pt plus 2pt minus 2pt}
% \begin{minipage}[t]{0.475\linewidth}
% \centering
% \footnotesize
% \resizebox{\linewidth}{!}{%
% \begin{tabular}{c|cc}
% \hline
% \textbf{Method} & \textbf{O-IoU} & \textbf{M-IoU} \\
% \hline
% LLaVA-7B~\cite{llava}          & 15.8 & 11.1 \\
% Chat-UniVi-7B~\cite{chatunivi} & 17.2 & 12.5 \\
% Qwen3-VL-8B~\cite{qwen3}       & 24.5 & 19.2 \\
% \hline
% \end{tabular}
% } % end resizebox
% \caption{Performance of pre-trained MLLMs in RVOS.}
% \label{ablation:MLLM}
% \end{minipage}
% \hfill
% \begin{minipage}[t]{0.475\linewidth}
% \centering
% \footnotesize
% \resizebox{\linewidth}{!}{%
% \begin{tabular}{c|ccc}
% \hline
% \textbf{Method} & \textbf{testA} & \textbf{testB} & \textbf{Parms(M)} \\
% \hline
% TRIS~\cite{TRIS}    & 32.4 & 29.6 & 52 \\
% PCNet~\cite{PCNet}  & 58.4 & 42.1 & 91 \\
% DViN~\cite{dvin}    & 63.8 & 57.0 & 54 \\
% WSRVOS               & 60.2 & 52.3 & 31 \\
% % TRIS~\cite{TRIS}    & 31.2 & 32.4 & 29.6 & 52 \\
% % PCNet~\cite{PCNet}  & 52.2 & 58.4 & 42.1 & 91 \\
% % DViN~\cite{dvin}    & 61.4 & 63.8 & 57.0 & 54 \\
% % WSRVOS              & 58.3 & 60.2 & 52.3 & 31 \\
% \hline
% \end{tabular}
% } % end resizebox
% \caption{Weakly-supervised RIS on RefCOCO (Mean IoU).}
% \label{ablation:ris}
% \end{minipage}

% \endgroup
% \vspace{-2mm} % 适当收紧表格与下文的垂直距离
% \end{table}

We show the qualitative results of WSRVOS across frames in Fig.~\ref{qualitative_results}. We can see it effectively identifies and segments target instances referred to by the text expressions. 

% \hl{Last, we report the model performance of WSRVOS in weakly-supervised referring image segmentation (RIS) task on RefCOCO dataset. As shown in Tab.~\ref{ablation:ris}, our WSRVOS outperforms several powerful weakly-supervised RIS framework, TRIS and PCNet, with a large margin on three different splits from RefCOCO~\cite{RIS1}.}

We compare the number of training parameters and inference speed 
% (measured by GFLOPs) 
across sota methods on Refer-YouTube-VOS. As shown in Tab.~\ref{tab:efficiency}, our WSRVOS achieves the best, \ie, it requires only 31M training parameters while running at 58 FPS, demonstrating its superior computational efficiency and real-time capability.

\subsection{Ablation studies}
% \vspace{1mm}
\label{sec:ablation}
We follow previous methods~\cite{soc,losh} to conduct ablation study on Refer-YouTube-VOS.

% \vspace{1mm}
\noindent\textbf{Contrastive referring expression augmentation} 
% \vspace{1mm}
% \noindent\emph{Effectiveness of CREA scheme.} 

\noindent
We introduce the contrastive referring expression augmentation (CREA) scheme to generate both positive and negative expressions. 
First, if we remove this scheme entirely and follow~\cite{TRIS} to use only the original expression as positive expression while sampling negative expressions from unrelated samples, as shown in line 1 in Tab.~\ref{tab:ablation-eg}, the performance drops significantly, with $\mathcal{J}$\&$\mathcal{F}$ decreasing from 36.5\% to 30.3\%. 
Second, if we exclude CREA for generating negative expressions and instead adopt negatives from other samples, the $\mathcal{J}$\&$\mathcal{F}$ decreases to 35.2\%.
Third, if we exclude CREA for generating positive expressions and instead simply duplicate the original expression to generate multiple positive expressions, we observe a performance degradation for $\mathcal{J}$\&$\mathcal{F}$ from 36.5\% to 33.7\%. 
These results demonstrate that our CREA scheme effectively enriches semantic details in positive expressions and introduces hard negative expressions, thereby enabling the model to learn more discriminative representations.

% . When the positive expression generation scheme is omitted, we obtain $P$ positive expressions by directly duplicating the original annotation. Likewise, in the absence of the negative expression generation scheme, we produce $N$ negative expressions by sampling descriptive texts from other videos. In the case where MLLM is only used to generate positive expressions, we observe 4.4\% and 4.7\% increases on Overall IoU and Mean IoU, respectively. On the other hand, if introducing only negative expressions, the model can still obtain a certain accuracy improvement ( \textit{e.g.}, 3.0\% Overall IoU and 3.1\% Mean IoU). Under the setting of introducing both positive and negative referring expressions, our WSRVOS obtains 5.9\% and 5.5\% higher Overall IoU and Mean IoU than the model without expression generation scheme. 

% \noindent\emph{Number of positive and negative expressions.} We vary the number of generated positive expressions $P$ from 2 to 10 and that of generated negative expressions $N$ from 12 to 60. The results are presented in Tab.~\ref{tab:ablation-PN}. We can observe that the model performs the best when $P=6$ and $N=48$, which are our default settings.
\begin{table}[tp]
\centering
\setlength{\tabcolsep}{0.7mm}
\setlength{\textfloatsep}{8pt plus 2pt minus 2pt}
\captionsetup{aboveskip=2pt, belowskip=0pt}
\begin{minipage}[t]{0.48\linewidth}  
\centering
\footnotesize
\begin{tabular}{cc|ccc}
\hline
\multicolumn{2}{c|}{\textbf{CREA}} & \multirow{2}{*}{$\mathcal J$} & \multirow{2}{*}{$\mathcal F$} & \multirow{2}{*}{$\mathcal{J}$\&$\mathcal{F}$} \\
\cline{1-2}
\textbf{pos.} & \textbf{neg.} &      &      \\
\hline
          &          & 27.2 & 33.5 & 30.3 \\
\ding{51} &          & 32.1 & 38.3 & 35.2\\
          & \ding{51} & 30.4 & 37.1 & 33.7 \\
\ding{51} & \ding{51} & \textbf{33.5} & \textbf{39.4} & \textbf{36.5} \\
\hline
\end{tabular}
% \caption{Ablation study on the contrastive referring expression augmentation scheme. pos. and neg. denote the positive and negative expression generation process, respectively.}
\caption{Ablation study on the contrastive referring expression augmentation scheme.}
\label{tab:ablation-eg}
\end{minipage}
\hfill
\setlength{\tabcolsep}{0.7mm}
\begin{minipage}[t]{0.48\linewidth}  
\centering
\footnotesize
\begin{tabular}{cc|ccc}
\hline
\multicolumn{2}{c|}{\textbf{V-L selection}} & \multirow{2}{*}{$\mathcal J$} & \multirow{2}{*}{$\mathcal F$} & \multirow{2}{*}{$\mathcal{J}$\&$\mathcal{F}$} \\
\cline{1-2}
\textbf{l2v.} & \textbf{v2l.} &      &      \\
\hline
          &          & 30.7 & 37.3 & 34.0 \\
\ding{51} &      & 32.2 & 38.5 & 35.3 \\
          &  \ding{51}  & 32.3 & 38.5 & 35.4 \\
\ding{51} & \ding{51} & \textbf{33.5} & \textbf{39.4} & \textbf{36.5} \\
\hline
\end{tabular}
% \caption{Ablation study on the bi-directional vision-language feature selection module. l2v. and v2l. denote the language-to-vision selection and the vision-to-language selection, respectively.}
\caption{Ablation study on the bi-directional vision-language feature selection module. }
\label{ablation:selection}
\end{minipage}
\vspace{-3mm}
\end{table}

% \vspace{1mm}
\noindent\textbf{Multimodal feature selection and interaction} 
% \vspace{1mm}

\noindent\emph{Effectiveness of bi-directional vision-language feature selection.} We introduce this module to bi-directionally select visual and linguistic tokens that are highly relevant to each other, enabling fine-grained multimodal alignment. To evaluate its effectiveness, we first propose a variant where we remove this module entirely. This variant (line 1 in Tab.~\ref{ablation:selection}) decreases $\mathcal{J}$\&$\mathcal{F}$ to 34.0\%. Second, we evaluate uni-directional variants that select only visual/linguistic tokens relevant to linguistic/visual ones. As shown in Tab.~\ref{ablation:selection}, these variants reduce the $\mathcal{J}$\&$\mathcal{F}$ from 36.5\% to 35.3\% and 35.4\%.

% Furthermore, we study the effect of the number of positive and negative referring expressions in Tab.~\ref{tab:ablation-PN}. No matter what value $N$ is set to, as $P$ increases, the model is able to utilize richer positive expressions to learn multimodal feature alignment, and reaches its peak when $P=6$. When $P$ is increased to 8, the segmentation accuracy decreases slightly. We suspect that using too many additional positive expressions generated by a MLLM will suppress original expression.
% On the other hand, when $N$ increases from 12 to 48, the accuracy of the model improves significantly, and when this number is increased to 60, the Overall IoU of the model begin to decrease slightly. This shows that although negative expressions can help improve the discriminative ability of the model, its number has to be balanced with that of positive expressions. 

\noindent\emph{Effectiveness of bi-directional vision-language feature interaction.} We enable bi-directional interaction between linguistic and visual features via cross-modal attention layers, enabling each modality to fully incorporate information from the other to enrich their own representation. As shown in Tab.~\ref{ablation:interaction}, removing this module entirely or disabling either direction of interaction leads to a performance degradation.

% To assess its effectiveness, we first evaluate a variant in which this module is entirely removed. As shown in line 1 of Tab.~\ref{ablation:interaction}, this results in a decrease of Overall IoU to 38.5\%. We further examine single-directional variants that select either visual features relevant to linguistic features (v2la.) or linguistic features relevant to visual features (l2va.). These variants reduce the O-IoU from 40.8\% to 39.5\% and 39.4\%, respectively, highlighting the superiority of the bi-directional V-L feature interaction module.

% \vspace{1mm}
\noindent\textbf{Instance-aware expression classification} 

% \vspace{1mm}
\noindent
Our proposed instance-aware expression classification (IEC) scheme perform proposal aggregation and expression matching to enhance the model in distinguishing positive from negative expressions. We evaluate its effectiveness in Tab.~\ref{ablation:components}. We can see removing this scheme (WSRVOS w/o IEC) leads to a notable performance drop, \ie, -10.6\% in $\mathcal J$ and -9.0\% in $\mathcal F$. 
Moreover, we introduce a variant of IEC scheme without performing proposal aggregation and expression matching, \ie, given proposal features $R$ corresponding to $P$ positive and $N$ negative expressions, we directly apply a linear classifier to $R$ for expression classification. 
%hereby guiding the model to distinguish between positive and negative expressions.
% encouraging feature similarities between proposals of positive expressions with positive expressions while suppressing those with negative expressions. 
We can see this variant (IEC w/o paem) decreases $\mathcal{J}$\&$\mathcal{F}$ from 36.5\% to 28.7\%. 
% Additionally, we evaluate another variant without applying the sigmoid activation and threshold filtering on the similarity map, \ie directly using the similarity map to achieve the proposal. This variant (IEC w/o bin) results in a performance drop in Mean IoU from 35.5\% to 33.8\%.  
Next, IEC scheme includes a classification flow and a segmentation flow to compute proposal-expression scores. Removing either flow (IEC w/o cls or IEC w/o smt) results in a performance decrease. Finally, we set the parameters of the classification flow using the transpose of linguistic features of expressions. If we instead follow the original MIL setting and use learnable parameters (\ie, cls$\rightarrow$cls-learn), the performance degrades.

\begin{table}[!t]
\centering
\begingroup
\captionsetup{aboveskip=2pt, belowskip=0pt}
\setlength{\textfloatsep}{4pt plus 1pt minus 1pt}
\setlength{\tabcolsep}{0.5mm}
\begin{minipage}[t]{0.45\linewidth}
\vspace{0pt}
\centering
\footnotesize
\begin{tabular}{cc|ccc}
\hline
\multicolumn{2}{c|}{\textbf{V-L interaction}} & \multirow{2}{*}{$\mathcal J$} & \multirow{2}{*}{$\mathcal F$}  & \multirow{2}{*}{$\mathcal{J}$\&$\mathcal{F}$} \\
\cline{1-2}
\textbf{l2va.} & \textbf{v2la.} &      &      \\
\hline
          &          & 31.2 & 37.9 & 34.5 \\
\ding{51} &          & 32.7 & 38.8 & 35.8 \\
          & \ding{51}& 32.7 & 38.7 & 35.7 \\
\ding{51} & \ding{51}& \textbf{33.5} & \textbf{39.4} & \textbf{36.5} \\
\hline
\end{tabular}
\caption{Ablation study on the bi-directional vision-language feature interaction module.}
\label{ablation:interaction}
\end{minipage}
\hfill
\begin{minipage}[t]{0.51\linewidth}
\vspace{0pt}
\centering
\footnotesize
\setlength{\tabcolsep}{0.4mm}
\begin{tabular}{p{2.5cm}|ccc}
\hline
\textbf{Method} & {$\mathcal J$} & {$\mathcal F$} & {$\mathcal{J}$\&$\mathcal{F}$}  \\
\hline
WSRVOS w/o IEC           & 22.9 & 30.4 & 26.6\\ 
IEC w/o paem             & 25.1 & 32.3 & 28.7 \\ 
IEC w/o cls              & 31.8 & 38.2 & 35.0 \\ 
IEC w/o smt              & 31.1 & 37.9 & 34.5 \\ 
cls$\rightarrow$cls-learn& 32.2 & 38.4 & 35.3 \\ 
% WSRVOS w/o PPF           & 25.5 & 32.5 & 29.0 \\ 
% PPF $\rightarrow$ PPF-single& 30.8 & 37.7 & 34.3 \\ 
% WSRVOS w/o TSR           & 29.3 & 34.9 & 32.1 \\ 
WSRVOS                   & \textbf{33.5} & \textbf{39.4} & \textbf{36.5} \\
\hline
\end{tabular}
\caption{Ablation study on the instance-aware expression classification scheme.}
\label{ablation:components}
\end{minipage}
\endgroup
% \vspace{-4mm}
\end{table}

\begin{table}[tp]
\centering
\captionsetup{aboveskip=2pt, belowskip=0pt}
\footnotesize
\setlength{\tabcolsep}{0.4mm}
\begin{tabular}{p{2.5cm}|ccc}
\hline
\textbf{Method} & {$\mathcal J$} & {$\mathcal F$} & {$\mathcal{J}$\&$\mathcal{F}$}  \\
\hline
% WSRVOS w/o IEC              & 22.9 & 30.4 & 26.6\\ 
% IEC w/o paem                & 25.1 & 32.3 & 28.7 \\ 
% IEC w/o cls                 & 31.8 & 38.2 & 35.0 \\ 
% IEC w/o smt                 & 31.1 & 37.9 & 34.5 \\ 
% cls$\rightarrow$cls-learn   & 32.2 & 38.4 & 35.3 \\ 
WSRVOS w/o PPF              & 25.5 & 32.5 & 29.0 \\ 
PPF $\rightarrow$ PPF-single& 30.8 & 37.7 & 34.3 \\ 
WSRVOS w/o TSR              & 29.3 & 34.9 & 32.1 \\ 
WSRVOS                      & \textbf{33.5} & \textbf{39.4} & \textbf{36.5} \\
\hline
\end{tabular}
\caption{Ablation study on the positive-prediction fusion strategy and temporal segment ranking constraint.}
\label{ablation:ppf}
\vspace{-4mm}
\end{table}

\begin{figure}[htbp]
    \centering
    \captionsetup{aboveskip=1pt, belowskip=0pt}
    \vspace{-2mm}
    \includegraphics[width=\linewidth]{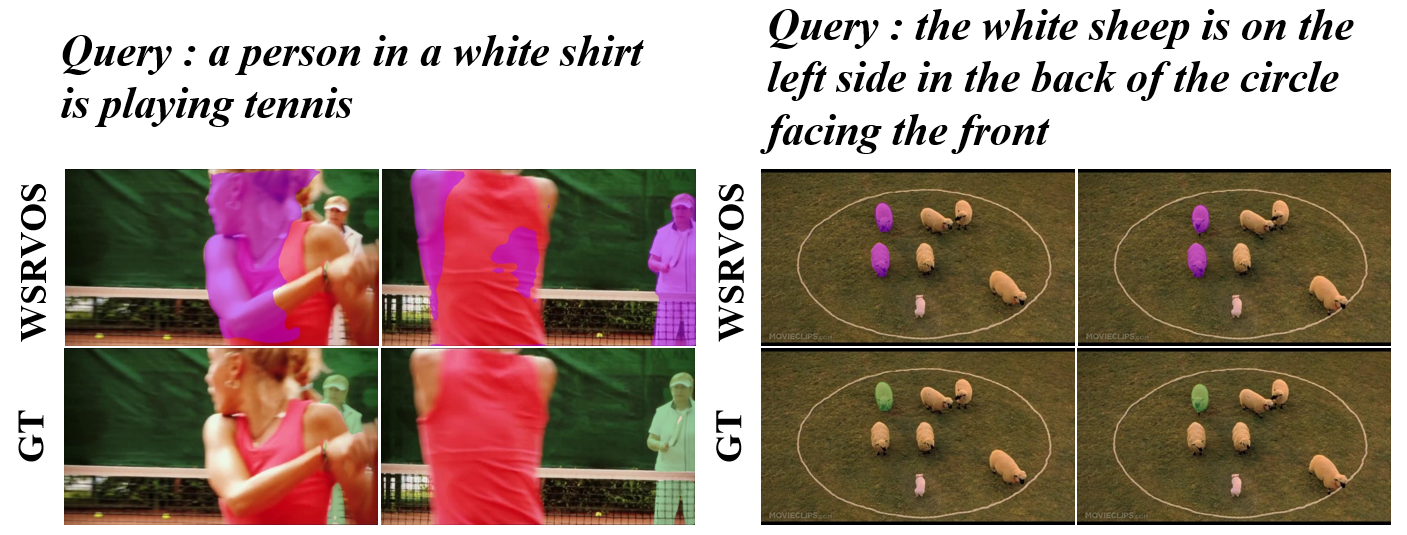}
    \caption{Failure cases. GT: ground truth.}
    \label{fig:tsr}
\vspace{-4mm}
\end{figure}

% \vspace{1mm}
\noindent\textbf{Positive-prediction fusion} 
% \vspace{1mm}

\noindent
% \textbf{Positive-prediction fusion strategy.}
% We propose PPF strategy to fuse the predictions from positive expressions to generate patch-wise pseudo-
% masks that serve as supervision signals. First, if we remove this strategy entirely, which means we don't generate pesudo-masks and don't utilize the segmentation loss, as shown in Tab.~\ref{ablation:segmentation}, there are -8.3\% and -8.1\% decreases on O-IoU and M-IoU. Second, if we utilize the segmentation loss but generate the pesudo-masks only utilizing one random positive expression, the M-IoU decreases to 33.3. These results collectively demonstrate the contribution of the fusion strategy to model performance. 
We propose positive-prediction fusion (PPF) strategy to fuse the predictions from multiple positive expressions to generate pseudo-masks that serve as supervision signals. As shown in Tab.~\ref{ablation:ppf}, completely removing this strategy (\ie, WSRVOS w/o PPF) leads to significant performance drops, \eg, -7.5\% in $\mathcal{J}$\&$\mathcal{F}$. Moreover, we provide a variant of PPF strategy in which the pseudo-mask for each frame is generated using only the prediction of a single randomly selected positive expression. This variant (\ie, PPF $\rightarrow$ PPF-single) shows a performance degradation of -2.7\% in $\mathcal J$, highlighting the effectiveness of using multiple positive expressions to generate robust pseudo-masks.

% \vspace{1mm}
\noindent\textbf{Temporal segment ranking} 
% \vspace{1mm}

\noindent
We propose temporal segment ranking (TSR) constraint to optimize the overlap values between mask predictions of temporally neighboring frames. As shown in Tab.~\ref{ablation:ppf}, removing this constraint (\ie, WSRVOS w/o TSR) leads to significant performance drops, \eg, -4.4\% in $\mathcal{J}$\&$\mathcal{F}$.
% We introduce a temporal consistency constraint module to enhance inter-frame temporal consistency of the generated pesudo-masks. As shown in Tab.~\ref{ablation:components}, completely removing this strategy (\ie, WSRVOS w/o TCC) leads to significant performance drops, \eg, -4.1\% in $\mathcal{J}$\&$\mathcal{F}$.
\noindent

\subsection{Failure cases}
We present some failure cases in Fig.\ref{fig:tsr} and summarize the key limitations. First, object occlusions can hinder segmentation. Second, when numerous instances of the same category appear simultaneously, WSRVOS may struggle to distinguish the target instance from surrounding ones.

\section{Conclusion}
% \vspace{2mm}
% \enlargethispage{-\baselineskip}
In this paper, we propose WSRVOS, a weakly-supervised referring video object segmentation framework using only text expressions as supervision. 
We first propose a contrastive referring expression augmentation scheme to generate positive and negative expressions via MLLMs. 
Next, we perform bi-directional vision-language feature selection and interaction on the visual and linguistic features. 
We then propose an instance-aware expression classification scheme to train the model to distinguish positive from negative expressions. Next, we introduce a positive-prediction fusion strategy to train our model with high-quality pseudo-masks. Finally, we design a temporal segment ranking constraint to encourage higher overlap among mask predictions from temporally closer frames than from distant ones. Extensive experiments on four benchmarks demonstrate the effectiveness of our WSRVOS.
\clearpage
\noindent\textbf{Acknowledgments.} This work was supported by the National Natural Science Foundation of China under Grant 62401393, the Fundamental Research Funds for the Central Universities, and the Xiaomi Young Scholar Project.

% \vspace{-1.0\baselineskip} 
{
    \small
    \setlength{\parskip}{0pt} 
    \bibliographystyle{ieeenat_fullname}
    \bibliography{main}
}

\clearpage
\setcounter{page}{1}
\setcounter{section}{0}
\setcounter{figure}{0}
\setcounter{table}{0}
\renewcommand{\thesection}{S\arabic{section}}
\renewcommand{\thefigure}{S\arabic{figure}}
\renewcommand{\thetable}{S\arabic{table}}
\maketitlesupplementary

\section*{Overview of supplementary material}
This supplementary material provides 1) details of datasets; 2) additional ablation studies on Refer-YouTube-VOS~\cite{Urvos}; 3) more visualization results.

\section{Details of Datasets}

A2D-Sentences~\cite{firstRVOS} contains 3,754 videos, split into 3,017 for training and 737 for testing. 
Each video is annotated with three or five frames, providing pixel-wise segmentation masks for various target instances. In total, the dataset includes 6,655 referring expressions, each of which corresponds to an instance within the annotated frames.
Refer-YouTube-VOS~\cite{Urvos} comprises 3,978 videos annotated with 15,009 referring expressions. Pixel-wise segmentation masks are provided for every fifth frame in these videos.  
Furthermore, following ~\cite{soc}, we evaluate the models trained on A2D-Sentences and Refer-YouTube-VOS by testing them on JHMDB-Sentences~\cite{firstRVOS} and Refer-DAVIS17~\cite{refdavis17} without finetuning, respectively. These two datasets extend the original vision-only datasets, JHMDB and DAVIS17, by incorporating rich text annotations. Specifically, JHMDB-Sentences is comprised of 928 videos, each paired with a referring expression. In contrast, Refer-DAVIS17 includes 90 videos and is annotated with a total of 1,544 referring expressions.

\section{Additional Ablation Studies}

\subsection{Contrastive referring expression augmentation} 
\begin{figure}[t]
  \centering
  \setlength{\abovecaptionskip}{0pt}
  \includegraphics[width=\linewidth]{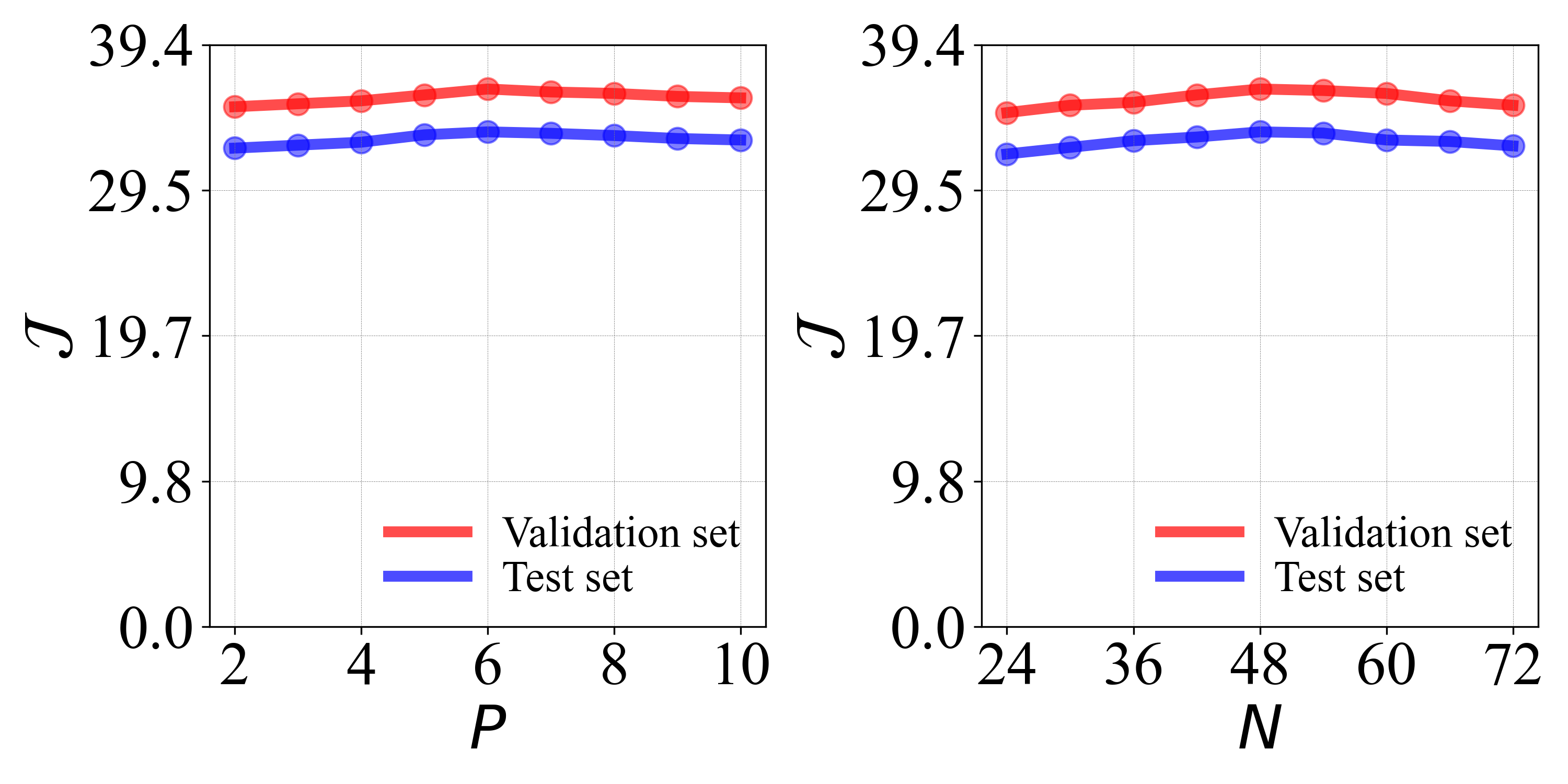}
  \caption{Parameter variation for the number of generated positive and negative expressions on the validation set and test set of Refer-YouTube-VOS, where the red line denotes the validation set and the blue line denotes the test set.}
  \label{fig:ablation-PN}
% \vspace{-5mm}
\end{figure}

\noindent\emph{Number of generated expressions.}
We vary the number of generated positive expressions $P$ from 2 to 10 and that of generated negative expressions $N$ from 24 to 72 on Refer-YouTube-VOS. The results on the validation set and test set are presented in Fig.~\ref{fig:ablation-PN}. Based on the validation set, we select $P=6$ and $N=48$ as our default settings, and this configuration also achieves the best performance on the test set.

\noindent\emph{Threshold in CREA.}
We vary the cosine similarity threshold in CREA from 0.7 to 0.9 on Refer-YouTube-VOS. The results on the validation set are presented in Tab.~\ref{ablation:crea}. It demonstrates that our default setting (WSRVOS w/ CREA(0.8)) performs the best.

% \noindent\emph{Number of generated expressions.}Tab.~\ref{ablation:crea} shows the results under different values of the cosine similarity threshold in CREA. It demonstrates that our default setting (WSRVOS(CREA(0.8))) performs the best.
% We can observe that the model performs the best when $P=6$ and $N=48$, which are our default settings.

% \noindent\emph{Different MLLMs for expression generation.}
% We further investigate the impact of different MLLMs for positive and negative expression generation. In Tab.~\ref{ablation:mllm} we denote  WSRVOS (VideoLLaMA3-7B) as a variant in which we leverage VideoLLaMA3-7B~\cite{videollama} for generating expressions. Other variants are denoted in a similar manner. We can see WSRVOS (VideoLLaMA3-7B) achieves very close performance to our default setting with Qwen3-VL-8B. Moreover, when using a larger MLLM, \ie, Qwen3-VL-30B, the performance can be slightly improved.  Since Qwen3-VL-8B is more deployable than Qwen3-VL-30B, we choose the former by default. 

% Qwen2.5-VL-7B~\cite{qwen2.5}  yields comparable results on all three metrics. Notably, adopting Qwen3-VL-30B~\cite{qwen3} leads to a slight performance improvement. However, considering the significantly higher computational cost, we employ Qwen3-VL-8B as our default MLLM for semantically abundant expression generation.

\begin{figure}[t]
  \centering
  \setlength{\abovecaptionskip}{0pt}
  \includegraphics[width=\linewidth]{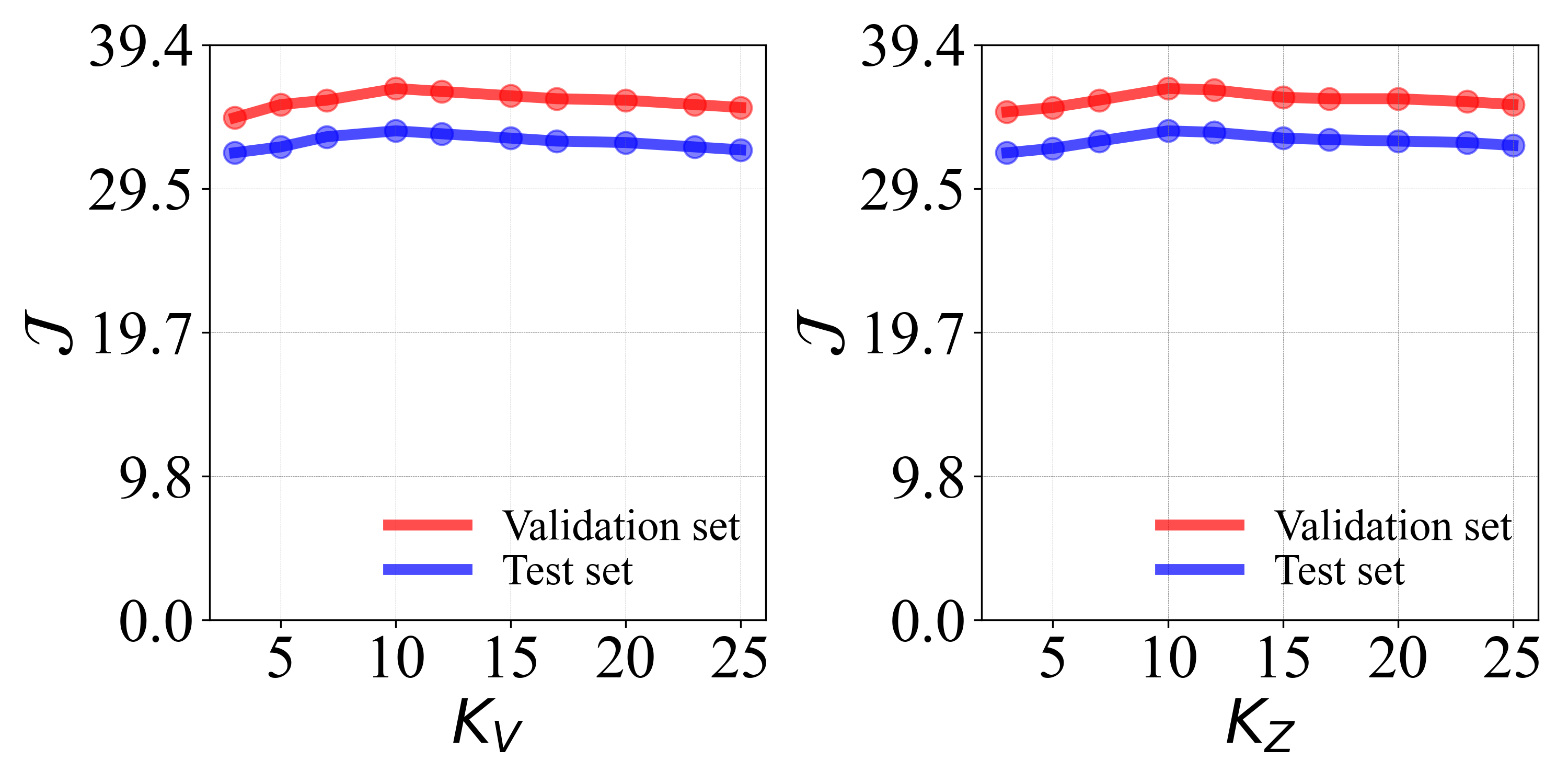}
  \caption{Parameter variation for the number of selected features in the bi-directional vision-language feature selection module on the validation set and test set of Refer-YouTube-VOS, where the red line denotes the validation set and the blue line denotes the test set.}
  \label{fig:ablation-K}
\vspace{-2mm}
\end{figure}

\begin{table}[th!]
\centering
% 关键修改：设置左对齐，并关闭单行标题强制居中
\captionsetup{justification=raggedright, singlelinecheck=false}
\footnotesize
\setlength{\tabcolsep}{0.4mm}
\begin{tabular}{p{4.2cm}|ccc}
\hline
\textbf{Method} & {$\mathcal J$} & {$\mathcal F$} & {$\mathcal{J}$\&$\mathcal{F}$} \\
\hline
% w/ full finetuning     & 33.2 & 39.2 & 36.2 \\ 
% w/ lora finetuning     & 33.4 & 39.2 & 36.3 \\ 
% WSRVOS w/ warm-up strategy    & 33.4 & 39.4 & 36.4 \\ 
% epoch 0 pseudo masks   & 16.4 & 19.2 & 17.8 \\ 
WSRVOS w/ CREA(0.7)           & 33.1 & 39.1 & 36.1 \\ 
WSRVOS w/ CREA(0.9)           & 33.2 & 39.2 & 36.2 \\ 
WSRVOS w/ CREA(0.8)           & \textbf{33.5} & \textbf{39.4} & \textbf{36.5} \\
\hline
\end{tabular}
\caption{Ablation study on the CREA module.}
\label{ablation:crea}
\vspace{-1mm}
\end{table}

\begin{figure*}
\begin{center}
\includegraphics[width=\linewidth]
{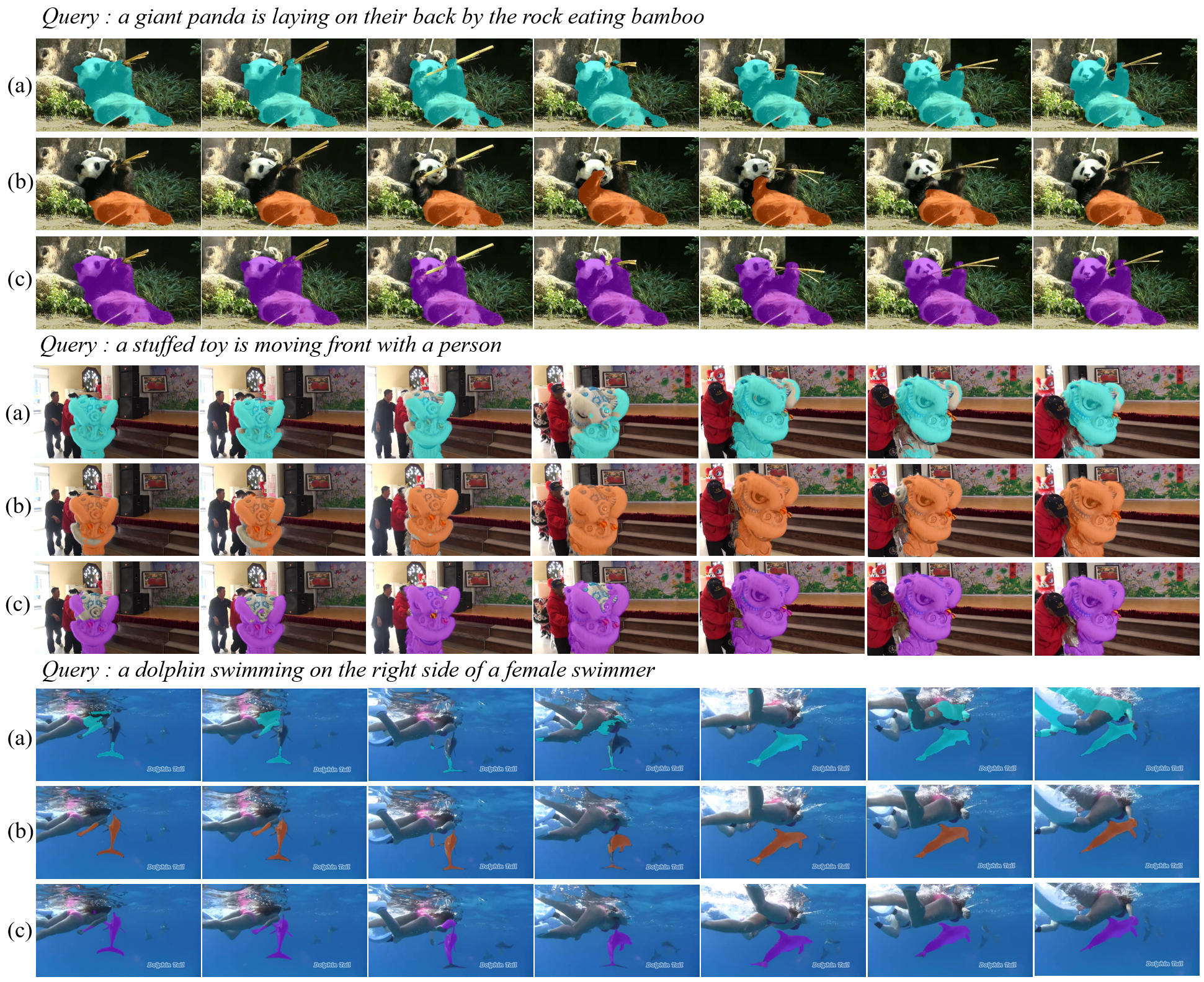}
\end{center}
   \caption{More visualization results on Ref-YouTube-VOS. (a) DViN (adapted weakly-supervised RIS method)~\cite{dvin}, (b) OCPG (point-supervised RVOS method)~\cite{ocpg}, (c) our proposed WSRVOS.}
\label{more vis}
 % \vspace{-3mm}
\end{figure*}

% \begin{table}[t]
% \footnotesize
% \setlength{\tabcolsep}{1.5mm}  
% \renewcommand{\arraystretch}{1.2}
% \centering
% \begin{tabular}{@{}l|ccc}
% \hline
% \textbf{Method} & $\mathcal{J}$ & $\mathcal{F}$ & $\mathcal{J}$\&$\mathcal{F}$ \\
% \hline
% % Video-LLaVA-7B~\cite{videollava} & 31.8 & 36.9 & 33.9 \\
% WSRVOS (VideoLLaMA3-7B)~\cite{videollama} & 33.2 & 38.8 & 36.0 \\
% % WSRVOS(Qwen2.5-VL-7B)~\cite{qwen2.5}      & 32.6 & 37.9 & 35.3 \\
% WSRVOS (Qwen3-VL-8B)~\cite{qwen3}  & 33.5 & 39.4 & 36.5 \\
% WSRVOS (Qwen3-VL-30B)~\cite{qwen3}  & 34.0 & 39.8 & 36.9 \\
% \hline
% \end{tabular}
% \caption{Comparison of different multimodal large language models for generating additional expressions on Refer-YouTube-VOS.}
% \label{ablation:mllm}
% \vspace{-3mm}
% \end{table}

% V-L selection(L-first)       & 33.3 & 29.2 & 31.2 \\
% V-L selection(V-L parallel)  & 33.1 & 29.1 & 31.1 \\

\subsection{Bi-directional vision-language feature selection}

% \noindent\emph{Order of bi-directional feature selection.}
% We investigate the impact of feature selection order in the bi-directional vision-language feature selection (V-L selection) module. In our framework, visual features are selected first. In Tab.~\ref{ablation:order} we introduce two variants: one selects linguistic features first, \ie {V-L selection(L-first)}, and another performs bi-directional feature selection in parallel, \ie {V-L selection(V-L parallel)}. Both variants performs slightly inferior to the original WSRVOS, attesting our choice. 

% \noindent\emph{Positive-only vision-language feature selection.} We further introduce a variant of the bi-directional vision-language feature selection module in which we select only the linguistic features relevant to visual features from positive expressions. As presented in Tab.~\ref{ablation:order}, this variant, \ie, {V-L selection (Positive-only)}, improves performance over the variant without the bi-directional vision-language feature selection module (\ie, WSRVOS w/o V-L selection), but still underperforms our full WSRVOS model. This is because negative expressions still exhibit partial similarities with the target instance, which can be leveraged for feature selection. 

\noindent\emph{Position of the bi-directional vision-language feature selection module.}
% We further compare different choices of transformer layers used for bi-directional vision-language feature selection. As shown in Tab.~\ref{ablation:order}, selecting features from the 4th, 8th, and 12th layers yields 36.3 in $\mathcal{J}$\&$\mathcal{F}$, leading to a minor decline.
Our proposed bi-directional vision-language feature selection module is integrated after the 3rd, 6th, 9th, and 12th layers of encoders. We also evaluate an alternative variant, denoted as V-L selection (All layers) in Tab.~\ref{ablation:order}, where the module is inserted after every encoder layer. However, this variant does not lead to clear performance improvement while substantially increasing the computational cost (training time increases from 12.3 hours to 14.1 hours and inference speed decreases from 58 FPS to 45 FPS).

% We additionally evaluate a variant, \ie {V-L selection(All layers)}, that selects all transformer layers for bi-directional vision-language feature selection. As shown in Tab.~\ref{ablation:order}, this configuration does not substantially improve performance while significantly increasing computational cost. Therefore, selecting a subset of representative layers remains a more efficient choice.

\noindent\emph{Number of selected features.}
Fig.~\ref{fig:ablation-K} presents the model performance under different numbers of selected visual and linguistic features, \ie $K_{V}$ and $K_{Z}$, on the validation set and the test set of Refer-YouTube-VOS. 
The results on the validation set indicates that $K_{V}=10$ and $K_{Z}=10$ achieve the best performance, and this setting consistently achieves the highest performance on the test set.
% which is our default setting.

\begin{table}[!t]
\centering
\footnotesize
\setlength{\tabcolsep}{1.0mm}  
\renewcommand{\arraystretch}{1.15}
\begin{tabular}{@{}l|ccc}
\hline
\textbf{Method} & $\mathcal{J}$ & $\mathcal{F}$ & $\mathcal{J}$\&$\mathcal{F}$  \\
\hline
WSRVOS w/o V-L selection      & 30.7 & 37.3 & 34.0    \\
% V-L selection (Positive-only) & 32.2 & 38.3 & 35.2   \\
V-L selection (All layers)    & 33.4 & \textbf{39.5} & \textbf{36.5} \\
WSRVOS                        & \textbf{33.5} & 39.4 & \textbf{36.5} \\
\hline
\end{tabular}
\caption{Ablation study on the bi-directional vision-language feature selection module.}
\label{ablation:order}
\vspace{-2mm}
\end{table}

\begin{table}[t!]
\centering
\footnotesize
\setlength{\tabcolsep}{0.4mm}
\begin{tabular}{p{2.5cm}|ccc}
\hline
\textbf{Method} & {$\mathcal J$} & {$\mathcal F$} & {$\mathcal{J}$\&$\mathcal{F}$}  \\
\hline
WSRVOS w/o TSR              & 29.3 & 34.9 & 32.1 \\ 
TSR $\rightarrow$ PSC & 32.6 & 38.7 & 35.6 \\ 
WSRVOS             & \textbf{33.5} & \textbf{39.4} & \textbf{36.5} \\
\hline
\end{tabular}
\caption{Ablation study on the temporal segment ranking constraint.}
\label{ablation:tsr}
\vspace{-4mm}
\end{table}

\subsection{Temporal segment ranking}
In Tab.~\ref{ablation:tsr} we introduce a variant in which we replace the temporal segment ranking (TSR) constraint with a pairwise segment consistency (PSC) loss, \ie, we simply encourage the IoU between the segmentation masks of each pair of consecutive frames to be as close to 1. We can observe that this variant (TSR $\rightarrow$ PSC) outperforms the variant without TSR (WSRVOS w/o TSR), but still performs inferior to our original WSRVOS with TSR.

\subsection{Warm-up strategy}
In Tab.~\ref{ablation:warm} we introduce a variant in which we apply a warm-up strategy during the early training stage, \ie, only the classification loss is optimized in the first 5 epochs, while the remaining losses are activated afterward. We can observe that this variant (WSRVOS w/ warm-up strategy) yields no performance gains. Therefore, no warm-up strategy is employed by default.

% \subsection{Warm-up strategy}
% In Tab.~\ref{ablation:warm} we introduce a variant in which we employ a warm-up strategy at the early stage of training, \ie, we simply apply only the classification loss for the first 5 epochs before introducing the full objective. We can observe that this variant (WSRVOS w/ warm-up strategy) yields no performance gains and performs slightly inferior to our original WSRVOS. Therefore, no warm-up strategy is employed by default.

% \subsection{Impact of warm-up strategy} No warm-up strategy is employed by default. We validate this by proposing a variant where only the classification loss is applied for the first 5 epochs of training. As shown in Tab.~\ref{ablation:warm}, this variant yields no performance gains.

\begin{table}[th!]
\centering
% 关键修改：设置左对齐，并关闭单行标题强制居中
\captionsetup{justification=raggedright, singlelinecheck=false}
\footnotesize
\setlength{\tabcolsep}{0.4mm}
\begin{tabular}{p{4.2cm}|ccc}
\hline
\textbf{Method} & {$\mathcal J$} & {$\mathcal F$} & {$\mathcal{J}$\&$\mathcal{F}$} \\
\hline
WSRVOS w/ warm-up strategy    & 33.4 & 39.4 & 36.4 \\ 
WSRVOS                 & \textbf{33.5} & \textbf{39.4} & \textbf{36.5} \\
\hline
\end{tabular}
\caption{Ablation study on the warm-up strategy.}
\label{ablation:warm}
\vspace{-1mm}
\end{table}

\section{More visualization results} We present additional qualitative results of WSRVOS and two comparison methods, \ie DViN (adapted weakly-supervised RIS method)~\cite{dvin} and OCPG (point-supervised RVOS method)~\cite{ocpg}), across video frames in Fig.~\ref{more vis}. 
% For comparison, we also report the visualization results from DViN~\cite{dvin} and OCPG(point-supervised method). 

% \newpage
% \bibliography{main}

% \end{document}

% \section{Rationale}
% \label{sec:rationale}
% % 
% Having the supplementary compiled together with the main paper means that:
% % 
% \begin{itemize}
% \item The supplementary can back-reference sections of the main paper, for example, we can refer to \cref{sec:intro};
% \item The main paper can forward reference sub-sections within the supplementary explicitly (e.g. referring to a particular experiment); 
% \item When submitted to arXiv, the supplementary will already included at the end of the paper.
% \end{itemize}
% % 
% To split the supplementary pages from the main paper, you can use \href{https://support.apple.com/en-ca/guide/preview/prvw11793/mac#:~:text=Delete%20a%20page%20from%20a,or%20choose%20Edit%20%3E%20Delete).}{Preview (on macOS)}, \href{https://www.adobe.com/acrobat/how-to/delete-pages-from-pdf.html#:~:text=Choose%20%E2%80%9CTools%E2%80%9D%20%3E%20%E2%80%9COrganize,or%20pages%20from%20the%20file.}{Adobe Acrobat} (on all OSs), as well as \href{https://superuser.com/questions/517986/is-it-possible-to-delete-some-pages-of-a-pdf-document}{command line tools}.

\end{document}